\documentclass[journal]{IEEEtran}

\usepackage{cite}
\usepackage[pdftex]{graphicx}
  
\usepackage{amsmath}
\usepackage{algorithmic}
\usepackage{array}

\usepackage{url}

\usepackage{color}
\usepackage{subfigure}
\usepackage{xcolor}

\usepackage{multirow}
\usepackage{amssymb,mathrsfs,amsmath}
\usepackage{algorithm}
\usepackage{algorithmic,eqparbox,array}

\newcommand{\ie}{{\em i.e.}}           % i.e.
\IEEEoverridecommandlockouts
\usepackage{graphicx}
\def\BibTeX{{\rm B\kern-.05em{\sc i\kern-.025em b}\kern-.08em
    T\kern-.1667em\lower.7ex\hbox{E}\kern-.125emX}}
\usepackage[font=small,labelfont=bf,tableposition=top]{caption}

\usepackage{blindtext}

\let\oldtwocolumn\twocolumn
\renewcommand\twocolumn[1][]{%
    \oldtwocolumn[{#1}{
\begin{center}
\vspace{-1cm}
    \includegraphics[width=\textwidth]{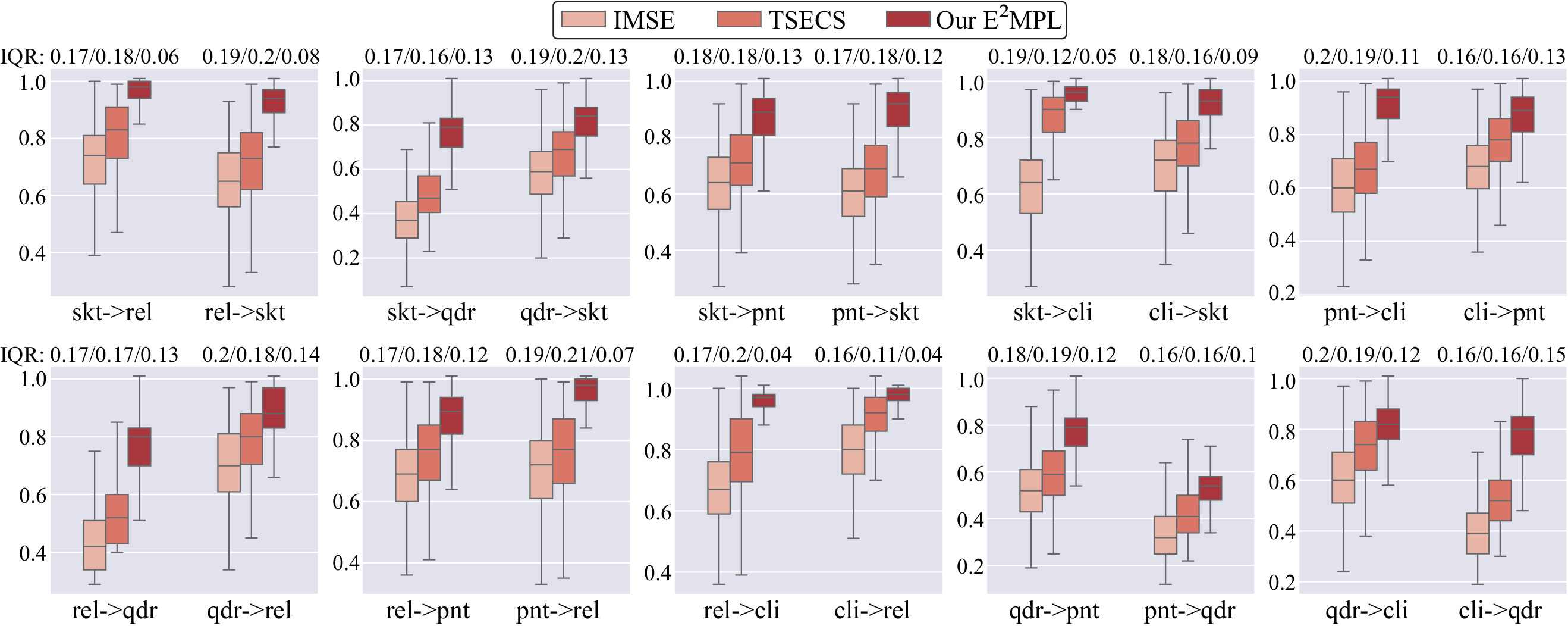}
    \captionof{figure}{Illustration of the accuracy boxplot and Interquartile range (IQR) of three FS-UDA models that adapt to 3600 test tasks with 5-way 1-shot setting. It is observed that our E$^2$MPL exhibits much more enduring and effective performance across diverse tasks in \emph{DomainNet} dataset.}
    \label{fig1}
\end{center}
    }]
}

\begin{document}

\title{E$^2$MPL: An Enduring and Efficient Meta Prompt Learning Framework for Few-shot Unsupervised Domain Adaptation}
\author{Wanqi~Yang, Haoran~Wang, Lei~Wang, Ge~Song, Ming~Yang, Yang~Gao
\thanks{Wanqi~Yang, Haoran~Wang, Ge~Song and Ming~Yang are with the School of Computer Science and Technology, and also with Ministry of Education Key Laboratory of NSLSCS, Nanjing Normal University, Nanjing, China. \protect (e-mail: yangwq@njnu.edu.cn, myang@njnu.edu.cn).}% <-this % stops a space
\thanks{Lei Wang is with the School of Computing and Information Technology, University of Wollongong, Australia. \protect (e-mail: leiw@uow.edu.au).}
\thanks{Yang Gao is with the Department of Computer Science and Technology, Nanjing University, Nanjing, China. \protect (e-mail: gaoy@nju.edu.cn).}}

% \begin{figure*}%{r}{0.32\textwidth}   0.8\textwidth
% \centering
% \includegraphics[width=2\columnwidth]{Fig_1.eps}%{1-shot.eps}
% \caption{Illustration the stability of three methods about few-shot unsupervised domain adaptation. Regardless of the 1-shot or 5-shot task settings, given a trained model, E$^2$MPL (our method) is more stable and effective over 3600 test tasks than some previous methods even if there are large domain gaps between the source and target domain data.}
% \label{fig1}
% %\vspace{-0.3cm}
% \end{figure*}

\maketitle

\begin{abstract}
Few-shot unsupervised domain adaptation (FS-UDA) leverages a limited amount of labeled data from a source domain to enable accurate classification in an unlabeled target domain. Despite recent advancements, current approaches of FS-UDA continue to confront a major challenge: models often demonstrate instability when adapted to new FS-UDA tasks and necessitate considerable time investment.
To address these challenges, we put forward a novel framework called Enduring and Efficient Meta-Prompt Learning (E$^2$MPL) for FS-UDA. Within this framework, we utilize the pre-trained CLIP model as the backbone of feature learning. Firstly, we design domain-shared prompts, consisting of virtual tokens, which primarily capture meta-knowledge from a wide range of meta-tasks to mitigate the domain gaps. Secondly, we develop a task prompt learning network that adaptively learns task-specific specific prompts with the goal of achieving fast and stable task generalization. Thirdly, we formulate the meta-prompt learning process as a bilevel optimization problem, consisting of (outer) meta-prompt learner and (inner) task-specific classifier and domain adapter. Also, the inner objective of each meta-task has the closed-form solution, which enables efficient prompt learning and adaptation to new tasks in a single step.
Extensive experimental studies demonstrate the promising performance of our framework in a domain adaptation benchmark dataset \emph{DomainNet}. Compared with state-of-the-art methods, our method has improved accuracy by at least 15.4\% and reduced the time by 68.5\% on average in 5-way 1-shot tasks, and improved accuracy by 8.7\% and reduced the time by 74.1\% on average in 5-way 5-shot tasks. Moreover, our approach exhibits more enduring performance than the other methods, \emph{i.e.,} being more stable across 3600 test tasks.

\end{abstract}


% \fi
\IEEEpeerreviewmaketitle

\section{Introduction}

In unsupervised domain adaptation (UDA), when the labeling cost is high or the access to labeled data is limited, it cannot be guaranteed that sufficient labeled data will be available for each category in the source domain. This limitation can severely impair the domain adaptation capabilities and consequently degrade the classification performance in the target domain. To address these issues, a setting known as few-shot unsupervised domain adaptation (FS-UDA) \cite{huang2021few}\cite{Yang}\cite{yu2023high} has emerged, which only leverages few-shot labeled data in the source domain for UDA, offering a potentially viable solution. An FS-UDA model is expected to acquire general knowledge from base classes during training and subsequently guide the classification of novel classes during testing. However, both insufficient labeled data in the source domain and large domain shifts between the source and target domains make FS-UDA a particularly challenging task.

% \begin{figure}%{r}{0.32\textwidth}
% \centering
% \includegraphics[width=0.98\columnwidth]{fig1up_new.pdf}
% \caption{Illustration of few-shot unsupervised domain adaptation in online shopping. Given few-shot images (\emph{e.g.}, 1-shot) (viewed as source domain), the task aims to train a classifier of products (\emph{e.g. TV, laptop, Kettle}) to identify the unlabeled images taken from buyers (viewed as target domain).}
% \label{fig1}
% %\vspace{-0.3cm}
% \end{figure}

Previous FS-UDA methods have addressed domain shifts primarily through the development of cross-domain similarity metrics \cite{huang2021few}, the extraction of high-level semantic features across domains \cite{yu2023high}, or the incorporation of domain adversarial loss into the meta-learning framework \cite{Yang}. However, we observed that these approaches show performance instability when adapting to various new tasks. %and generally require extensive time commitments.
As illustrated in Fig. \ref{fig1}, compared to our E$^2$MPL, the accuracy boxplots of previous methods (IMSE \cite{huang2021few} and TSECS \cite{yu2023high}), tested on 3600 distinct tasks, perform more dispersed with a lower median and a higher Interquartile range (IQR) \footnote{A box plot is a method for demonstrating graphically the locality, spread and skewness groups of numerical data through their quartiles, where Median (Q2) is the middle value, and Interquartile range (IQR) is the distance between the upper and lower quartiles, \ie, IQR=Q3-Q1, for statistical dispersion.}. %This indicates that they suffer from poor stability and unsatisfied average performance across different tasks. 

\emph{We believe that a promising FS-UDA model should be able to i) generalize effectively and stably to new and unseen tasks, and ii) ensure that the learning process is computationally efficient while maintaining high performance.} %Emphasizing these critical capabilities, it is essential that the model maintains high performance while requiring minimal training time.

To achieve this goal, we employ the pre-trained CLIP as the backbone of our model and introduce learnable virtual prompts as the inputs to CLIP for a lightweight yet promising update, enabling rapid adaptation to downstream various tasks without necessitating full parameter updates of CLIP. 
However, conventional visual prompt learning methods \cite{vpt}\cite{AD-CLIP} can easily overfit a limited number of training samples, as well as hardly address the domain shift. This could hinder the generalizability of the CLIP models for FS-UDA tasks. In light of the generalizability challenges faced by prompt learning, meta-prompt learning emerges as a viable alternative. Several meta prompt learning methods \cite{DAM-VP}\cite{GRAM}\cite{ProMetaR} leverage the meta learning strategy to learn the general prompts. For instance, DAM-VP \cite{DAM-VP} introduces a specific meta-prompt vector for each data subset, tailored to visual differences to mitigate the variations of data distribution within the same dataset, but it is not suitable for substantial discrepancy between domains. Prompt Meta-Regularization (ProMetaR) \cite{ProMetaR} was proposed to employ meta-regularization to alleviate overfitting, but it is limited by its reliance on text prompts and learning prompts is relatively time-consuming, requiring multiple updates for each task. Consequently, there is an urgent need to develop a more effective and efficient meta-prompt learning framework specifically designed for FS-UDA tasks.

To facilitate domain adaptation and task generalization, we design domain-shared prompts and task-specific prompts. Specifically, we embed domain-shared prompts into the feature embedding module to learn the general features across domains. To better adapt CLIP to various tasks, we make additional use of the pre-trained prompt network to generate task-specific prompts, which not only mitigates domain gaps but also helps the model focus on crucial information for the current task, resulting in improved stability. %and robustness.
%For the domain shift, we embed the domain-shared prompts into the feature embedding module to learn the general features across domains. Also, in order to better adapt CLIP to various tasks, we make additional use of the \re{pre-trained} prompt network to generate task-specific prompts, which can improve the performance stability. 
%As for meta training, we first collect an auxiliary data set that consists of a labeled source domain and an unlabeled target domain, which share a sufficient number of base classes. From the auxiliary data set, we randomly sample a collection of few-shot UDA tasks as in \cite{Yang}. With this collection of meta tasks, we perform episodic training for domain adaptation and classification. Through this process, we expect to learn meta prompts based on pre-trained CLIP to quickly adapt to downstream tasks. %\re{with MLP}

Therefore, we propose a novel Enduring and Efficient Meta-Prompt Learning Framework for FS-UDA, namely E$^2$MPL. Formally, the proposed meta-prompt learning process is formulated as a bilevel optimization problem consisting of the (outer) meta-prompt learner and the (inner) base learner \cite{DBLP:conf/icml/FranceschiFSGP18}. We design the meta-prompt learner over the collection of meta-tasks to obtain the domain-shared prompts and task-specific prompts aforementioned. Meanwhile, we build two base learners (\emph{i.e.,} a classifier and a domain adapter) for each of these meta-tasks, respectively. The classifier learns the optimal classification parameter for this task, while the domain adapter learns the optimal projection parameter to transform the target domain data close to the source domain data. To make the proposed meta-prompt learning process computationally efficient, instead of the time-consuming iterative calculation, both the classifier and the adapter enjoy a globally optimal closed-form solution that can be calculated in one step within each task. Thus, our E$^2$MPL can efficiently adapt the model to new tasks. Our main contributions can be summarized as follows. 
\begin{enumerate}
\item \textbf{A novel meta-prompt learning framework E$^2$MPL for FS-UDA}. 
We propose a bilevel optimization meta-prompt learning framework, where the outer meta-prompt learner captures meta-prompts from meta-tasks and the inner classifier and domain adapter are designed for each task with closed-form solutions, enabling efficient and stable adaptation to new tasks.
%By designing closed-form solutions for the base learner for each task, our E$^2$MPL can be efficiently trained and adapted to new tasks.

\item \textbf{Domain-shared and task-specific prompts.} Leveraging CLIP’s pretrained vision backbone, we introduce domain-shared and task-specific prompts optimized through meta-learning, which collaboratively enhance cross-domain alignment while preserving task-specific discriminability.

\item \textbf{Enduring and efficient performance.} Extensive experiments on \emph{DomainNet} validate our EMPL  achieves more effective and enduring performance across diverse tasks, and takes significantly less time to adapt to new tasks, compared to the alternative. %We hope that our work could serve as a quality baseline for further research along this line.
\end{enumerate}

% \begin{table*}
% \centering
% %\small
% \caption{The main differences between the cross-domain FSL methods and our MELDA.}\label{tab:table1}
% %\vspace{-0.3cm}
% \renewcommand{\arraystretch}{1.2}
% \setlength{\tabcolsep}{7pt}
% \begin{tabular}{cccc}\hline
% \textbf{Methods} & \textbf{Domain adaptation} & \textbf{New tasks} & \textbf{Label available in target domain?}\\\hline
% \textbf{TML} \cite{DBLP:conf/uai/KangF18} & between meta tasks and target domain data & FSL in target domain &  Yes\\
% \textbf{MLDA} \cite{sahoo2018meta}  & between meta tasks and new tasks & FSL in target domain & Yes \\
% %\textbf{DSN \cite{simon2020adaptive}}&
% \textbf{LFT} \cite{tseng2020cross} & between meta tasks in different domains & FSL in target domain & Yes\\
% %\textbf{MFR} \cite{guo2020learning} & transfer between meta-tasks & FSL in target domain & Yes \\
% \textbf{MELDA} for FS-UDA (ours)  & within every meta task & Few-shot UDA & \textbf{No} \\\hline
% \end{tabular}
% %\vspace{-0.3cm}
% \end{table*}

\section{Related Work}\label{relatedwork}

\subsection{Unsupervised Domain Adaptation} 
The UDA setting aims to reduce the domain gap and leverage sufficient data from the labeled source domain to achieve classification in the unlabeled target domain. Many UDA methods \cite{DBLP:journals/corr/TzengHZSD14,DBLP:journals/corr/Long015,DBLP:conf/cvpr/SaitoWUH18} are based on the maximum mean discrepancy to minimize the difference in features between domains. %They have shown effective performance on domain adaptation. 
By constructing deep networks, several methods \cite{long2016unsupervised,long2017deep,pan2019transferrable} have learned the domain-invariant representation, which is transferable between different domains. Long \emph{et al.} \cite{DBLP:journals/corr/Long015} introduced multiple domain adaptation modules in the high layers of deep convolutional network to match the mean embeddings of the distributions according to the maximum mean discrepancy criterion. Subsequently, they \cite{long2017deep} proposed a joint maximum mean discrepancy criterion to align the distributions of multiple domain-specific fully connected layers. Roy \emph{et al.} \cite{roy2019unsupervised} developed a unified deep domain adaptation framework and built domain alignment layers to match the feature distributions between different domains. If there are pseudo-labels with high confidence, UDA is converted to SSDA, which is beneficial for semi-supervised learning effects. By applying a gradient-variance-based selection mechanism, Yang \emph{et al.} \cite{y4} exploits a friendly subset instead of the entire open-set dataset to enhance the ID classification capacity of the model, which can be applied to the SSDA problem.

Moreover, adversarial training is widely used to tackle domain shifts. %This category of methods applied GANs to reduce the domain gap. 
There are several methods \cite{DBLP:conf/icml/GaninL15,DBLP:conf/cvpr/TzengHSD17,chen2019progressive} that developed domain-invariant feature generators and a domain discriminator to distinguish their authenticity/fakeness. DANN \cite{DBLP:conf/icml/GaninL15} learned domain-invariant features by training a domain classifier with a gradient reversal layer. ADDA \cite{DBLP:conf/cvpr/TzengHSD17} provided a generalized framework to combine adversarial learning, discriminative feature learning, and untied weight sharing. %They performed domain adaptation by using an adversarial loss and learning two domain-specific encoders.
CDAN \cite{long2018conditional} used discriminative classification predictions to align the domains. %CDAN \cite{long2018conditional} built a multilinear map of the high-level feature representation and classification prediction between the source and target domains to adapt the map between them. 
MCD \cite{DBLP:conf/cvpr/SaitoWUH18} maximized the discrepancy between task-specific classifiers to perform adversarial learning with the feature generator. To learn domain-invariant and semantic representations, a graph convolutional adversarial network \cite{ma2019gcan} was built to jointly perform the alignment of the data structure, domain and class centroid. In addition, contrast learning is also commonly used in UDA. CPRC \cite{y1} generates captions directly from images using the automatically learned cross-modal generator. For an unseen class from the UDA setting, CRV \cite{y3} holds a realistic setting that unlabeled data may come from unseen classes in the labeled set. CKGE \cite{y5} provides explainable recommendations with the consideration of different learning motivations from talents. In sum, existing UDA methods achieved domain adaptation with sufficient labeled source domain data. However, they would not work when encountering the issues of scarce labeled source domain and task-level generalization that exist in our FS-UDA setting. 
% \textbf{Adversarial Image Translation Methods.}
% This category of methods \cite{sankaranarayanan2018generate,hu2018duplex,fu2019geometry} conducted adversarial image translation for domain adaptation by generating the source-like or target-like images from real images. Coupled GAN (CoGAN) \cite{liu2016coupled} trained a coupled generative model to learn the joint data distributions of the two domains. UNIT \cite{liu2017unsupervised} extended CoGAN to learn image translation by integrating variational autoencoder and GAN. ADGAN \cite{sankaranarayanan2018generate} simultaneously performed image translation and learned the shared feature embedding, by imposing the source and target distributions to be close in a latent space. SBADA \cite{russo2018source} employed the structure of CycleGAN for image translation and proposed target self-labeling and class consistence between source domain images and their circle generation results.
\subsection{Prompt Learning for Computer Vision}
% Currently, prompt learning has recently made its way into the computer vision. The main goal of prompt learning is to leverage pre-trained language models to provide valuable knowledge for downstream tasks through text prompts. Concretely, CoOp\cite{CoOp} improves few-shot image classification by optimizing continuous prompts to fine-tune CLIP. CoCoOp\cite{CoCoOp} proposes learning conditional prompts based on image features to further improve the generalizability of CoOp. Then, APPLeNet\cite{APPLeNet} argues that the potential of VLMs for generalization task in remote sensing has not been fully realized, it introduce an attention-driven injection module to generate visual tokens from visual content features and style properties. Moreover, AD-CLIP\cite{AD-CLIP} thinks the above methods do not take into account domain gaps, so it conditions prompt learning on image style and content features simultaneously to learn domain-invariant and class-generalizable knowledge. However, these methods do not take into account scenarios where is no intersection between the training class and the test class.
Currently, prompt learning has recently been integrated into the field of computer vision. The primary objective of prompt learning is to utilize pre-trained models to offer valuable insights for downstream tasks through visual prompts. Concretely, learning methods for single-modal visual prompt learning include concatenating optimizable vector sequences \cite{vpt}\cite{L2P}\cite{CSVPT}\cite{ProD}, adding pixel-level optimizable disturbance \cite{VP, EVP, C-AVP}, learning prompt network layer \cite{Pro-Tuning, PGN, LION}, component-oriented combinatorial prompt learning \cite{DAM-VP}, network structure search \cite{NOAH}, \emph{etc}. Concatenated optimizeable vector sequences based on the Transformer structure are generated by concatenating additional optimizable vector sequences, such as VPT \cite{vpt}, on top of the original input sequence or each layer feature sequence of the Transformer structure. The inclusion of pixel level optimization allows for the disturbance independent of the model structure, enabling direct addition of an optimized random disturbance block or a rectangular box to the pixel space of the input image, such as VP \cite{VP}. The learning prompt network layer serves as a plug-in prompt module added primarily between the backbone network layers or as a generative prompt module outside the backbone network, such as Pro-Tuning \cite{Pro-Tuning}, PGN \cite{PGN}. Composition prompt learning for specific components involves designing different prompt templates for various data categories, such as DAM-VP \cite{DAM-VP}. The network structure search involves selecting a parameter-effective method at random for different downstream datasets to tune and then choosing the best performing one as the final prompt on that dataset. For example, NOAH \cite{NOAH} integrates the adapter \cite{Adapter}, LoRA \cite{LoRA}, and VPT \cite{vpt} as a splicable prompt module. %\ywq{However, some visual prompt-tuning methods do not account for the large domain gaps between the source domain training sets and the target domain test sets, nor do they consider generalization to new classes during testing.}

There are also multimodal prompt learning methods. CoOp \cite{CoOp} enhances the classification of few-shot images by optimizing continuous prompts to fine-tune CLIP. CoCoOp \cite{CoCoOp} proposes learning conditional prompts based on image features to further improve the generalization of CoOp. Moreover, APPLeNet \cite{APPLeNet} argues that the potential of Visual Language Model in remote sensing generalization task has not been fully realized; it introduces an attention-driven injection module to generate visual tokens from visual content features and style properties. Additionally, AD-CLIP \cite{AD-CLIP} argues that the aforementioned methods fail to take domain gaps into account. Therefore, it conditions prompt learning of both image style and content features simultaneously in order to acquire domain-invariant and class-generalizable knowledge. ProD \cite{ProD} utilized the prompting mechanism in the transformer to disentangle the domain-general and domain-specific knowledge from the backbone features for cross-domain few-shot image classification. However, these methods for prompting on text encoders are usually limited by text input. In our work, we need useful prompts to deal with domain gap and task generalization, as well as an efficient prompt  learning method.  % we designed is mainly inspired by VPT \cite{vpt} and PGN \cite{PGN}, which both store the ability of meta-knowledge to achieve category generalization and alleviate domain gaps.}
% \re{VPT\cite{vpt} is another visual prompt learning, it does not need to select text to as prompts, while set up learnable virtual tokens to achieve the purpose of fine-tuning ViT. ProD\cite{ProD} proposed a visual prompt to alleviate domain gaps for cross-domain few-shot learning, it concatenates a DG and a DS prompt to the backbone feature. The DG prompt is learnable and shared by all the training domains, while the DS prompt is generated from the domain-of-interest on the fly. DAM-VP\cite{DAM-VP} considered that the distribution of image datasets can vary greatly, and using a generic prompt specific to each dataset may not effectively address the complex distribution shift, so it cluster the downstream dataset into subset, and each subset has its own prompt optimized separately, which called meta-prompt. However, the generalization and representation ability of virtual tokens in these methods is too weak.}

\subsection{Meta-Learning for Few-shot Learning}
%In few-shot learning (FSL), a $N$-way and $K$-shot task leverages $K$ labeled samples per category to train a classifier to classify the $N$ categories. In recent years, meta-learning has gained significant attention and has been widely applied to FSL tasks. The existing meta-learning methods used for FSL can be primarily divided into two aspects: optimization-based methods and metric-based methods. 
In few-shot learning (FSL), it is key to leverage auxiliary data to train a generalizable model to prevent overfitting, even if only a few samples per category are available for the current task. In recent years, common few-shot learning methods can be mainly divided into two aspects: meta-learning optimization-based methods and metric-based methods.

Meta-learning optimization-based methods \cite{DBLP:conf/iclr/RaviL17,DBLP:conf/icml/FinnAL17} typically involve training a meta-learner on a collection of meta-tasks to learn general model parameters, \emph{i.e.,} initializing the parameters and hyper-parameters that can adapt to new tasks. For example, MAML \cite{DBLP:conf/icml/FinnAL17} and Meta-SGD \cite{DBLP:journals/corr/LiZCL17} learned a well-initialized model that updated the direction and learning rate, respectively. LEO \cite{DBLP:journals/corr/abs-1807-05960} extended MAML to learn a latent representation space. In addition, Franceschi \emph{et al.} formulated meta-learning through bilevel optimization \cite{DBLP:conf/icml/FranceschiFSGP18}, while Qin \emph{et al.} proposed the concept of intra-domain and inter-domain meta-knowledge \cite{Qin}, which also inspired us. %Our method is related to optimization-based methods. 

\begin{figure*}[t]
\centering
\includegraphics[width=2\columnwidth]{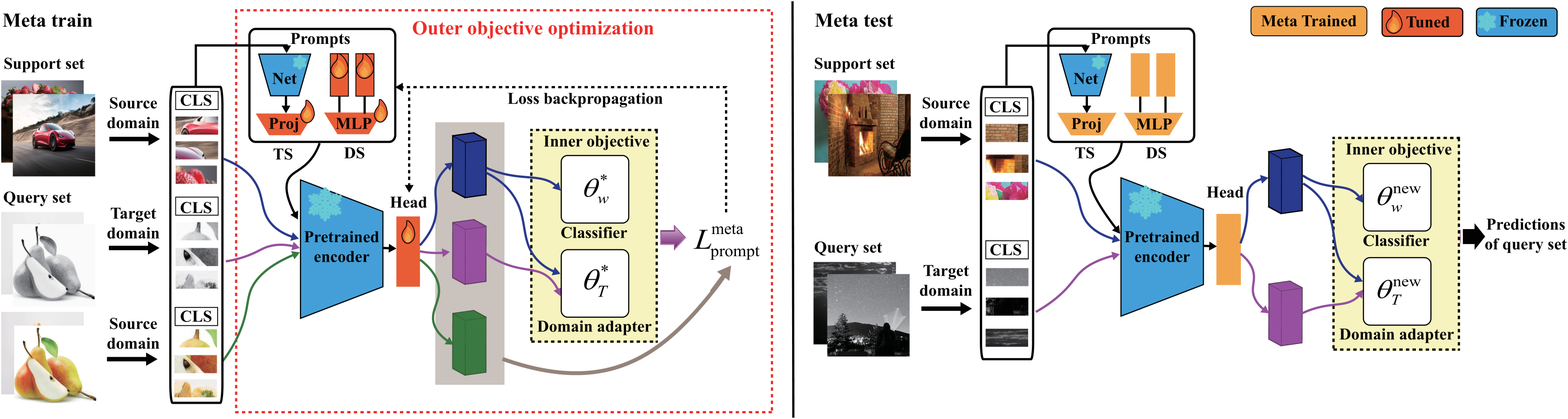}
%\vspace{-0.1cm}
\caption{Illustration of the proposed E$^2$MPL for FS-UDA. For each meta task, the support set, query set, domain-shared prompt (DS), and task-specific prompt (TS) are first fed to the pretrained CLIP image encoder and head layer to train a classifier and a domain adapter to get the optimal parameters $\theta_\omega^*$ and $\theta_\mathcal{T}^*$, respectively. Then, the query set from both source and target domains is used to calculate the meta-train loss, including the source domain classification loss $\mathcal{L}^{\text{meta}}_c$, target domain entropy minimum loss $\mathcal{L}^{\text{meta}}_d$, and class discrimination loss $\mathcal{L}^{\text{meta}}_f$. Finally, these losses are back-propagated to update the prompt module and head layer. The above process is conducted repeatedly over all meta tasks. For a test task, the support set is fed to the trained model to get the optimal $\theta_\omega^{\text{new}}$ and $\theta_\mathcal{T}^{\text{new}}$ for final prediction in the query set.} % and $\mathcal{L}^{\text{meta}}_d$, respectively\emph{at the top},\emph{At the bottom}, 
\label{fig2}
%\vspace{-0.4cm}
\end{figure*}

On the other hand, metric-based methods \cite{DBLP:journals/corr/SnellSZ17, li2019revisiting, FEAT} focus on learning a general feature metric space through episodic training on the auxiliary dataset \cite{DBLP:journals/corr/VinyalsBLKW16}. Classically, ProtoNet \cite{DBLP:journals/corr/SnellSZ17} learns the class prototypes in the support set and then classifies the query samples according to their maximum similarity to these prototypes. Li \emph{et al.} meanwhile leveraged the covariance matrix \cite{li2019revisiting} and local descriptors \cite{DBLP:conf/aaai/LiXHWGL19} for the measurement of image to class. 
Note that in the above methods, the support and query sets in an FSL task are usually in the same domain. They are not capable enough to handle the domain gap between the support set in source domain and the query set in target domain that exists in our FS-UDA setting.

\section{Methodology}
In this section, we begin by presenting the definition of FS-UDA problem. Next, we detail the design of our prompt module. Subsequently, we introduce our meta-prompt learning framework grounded in bilevel optimization \cite{DBLP:conf/icml/FranceschiFSGP18}. Lastly, we detail the implementation of the classifier and domain adapter, respectively.

\subsection{Problem Definition}
\textbf{A \emph{N-way, K-shot} UDA task.} The FS-UDA setting involves two domains in total: a source domain $\mathcal{S}$ and a target domain $\mathcal{T}$, both of which are shared by all tasks. A \emph{N-way, K-shot} UDA task consists of a support set $D_{\mathcal{S}}^\text{sup}$ derived from $\mathcal{S}$ and a query set $D_{
\mathcal{T}}^\text{que}$ derived from $\mathcal{T}$. The support set $D_{\mathcal{S}}^\text{sup}$ comprises $N$ classes with $K$ samples per class from the source domain. Meanwhile, the query set $D_{\mathcal{T}}^\text{que}$ contains $N\times N_{q}$ target domain samples from the same set of $N$ classes as the support set. Our goal is to train a general model capable of quickly generalizing to new tasks, accurately classifying the query set $D_{\mathcal{T}}^\text{que}$ given the few-shot support set $D_{\mathcal{S}}^\text{sup}$ of the new task. 
Note that the classes in the meta-testing tasks are not observed during the meta-training, ensuring that the model could generalize to previously unseen classes during the meta-testing. 

\textbf{Model architecture.}
Our framework mainly consists of a pre-trained CLIP image encoder $f_e$, a head layer $h_v$, a task-specific classifier, a task-specific adapter and a learnable prompt module. The pretrained CLIP encoder is frozen during training, and the classifier and adapter can directly calculate their parameters without optimization; Thus, only the prompt module and head layer need to be optimized.

\textbf{The flowchart of our framework.} 
Figure \ref{fig2} illustrates the meta-train and meta-test process of our E$^2$MPL. During the meta-train phase, in each episode, we construct a support set $D_{\mathcal{S}}^\text{sup}$ from the source domain, a query set $D_{\mathcal{T}}^\text{que}$ from the target domain, as well as an additional query set $D_{\mathcal{S}}^\text{que}$ from source domain for classifier training. \emph{For each meta-task}, the support set, query set, domain-shared prompts and task-specific prompts are first passed through the pre-trained encoder and the head layer to obtain the features of the current task, and then sent to the classifier and domain adapter to calculate their optimal parameters $\theta_\omega^*$ and $\theta_\mathcal{T}^*$, respectively. Afterwards, the query set ($D_{\mathcal{S}}^\text{que}$ and $D_{\mathcal{T}}^\text{que}$) is used to calculate the meta-prompt loss (\emph{i.e.,} classification loss $\mathcal{L}^{\text{meta}}_c$, entropy minimum loss $\mathcal{L}^{\text{meta}}_d$, and class discrimination loss $\mathcal{L}^{\text{meta}}_f$) among the meta-train tasks. 
Finally, the meta-prompt loss is backpropagated to optimize the prompt parameters and the head-layer parameter. 
During the meta-test phase, these prompt parameters are not updated. \emph{For each test task}, the support and query sets are fed into the classifier and adapter to calculate their optimal parameters $\theta_{\omega}^{\text{new}}$ and $\theta_\mathcal{T}^{\text{new}}$. Based on these parameters, the query set from the target domain is projected into the source domain space for the final prediction. Finally, we calculate the average classification accuracy of different test tasks for performance evaluation.

\subsection{Meta-Prompt Design}\label{dspl}
In the setting of FS-UDA, it is not feasible to train a model with a large number of parameters relying solely on a small number of data samples. Therefore, we employ the pre-trained CLIP as the backbone and introduce meta-prompt learning for a lightweight update. Meta-prompt learning enables the model to acquire meta-knowledge from multiple meta-tasks, which can achieve the domain adaptation and task generalization. To this end, we design domain-shared and task-specific prompts.

\textbf{Domain-shared prompts}. Existing prompt learning methods usually align the prompts between the source and target domains or the image and text prompts to mitigate domain gaps \cite{AD-CLIP}, but they require a large amount of data support to enhance the effectiveness of domain alignment. In light of this issue and inspired by VPT \cite{vpt}, we design domain-shared prompt tokens between the source and target domains that are concatenated by feature embedding for model training. 
%\re{concatenated by both the source and target domains} 
This approach encourages the model to learn similar feature representations between the source and target domains to reduce the domain gap. %Also, we observed that the variability and value of multiple trained tokens is relatively small, and thus we add an MLP layer for the tokens before feeding them to the network.

\textbf{Task-specific prompts}. Existing pre-trained models (such as CLIP) usually have poor generalization ability for unseen classes after prompt tuning, since they suffer from overfitting to the training classes \cite{CoCoOp}. Drawing inspiration from PGN \cite{PGN} and aiming to better adapt the model to new classes, we introduce prior knowledge by utilizing a pre-trained prompt network that is frozen to preserve the class generalization of the model. The prompt network can generate input-dependent prompts for the current task, allowing each image to receive specific prompts instead of using a fixed set of prompts, enabling the model to quickly adapt to unseen classes.

Specifically, the design approach for inputting domain-shared prompts and task-specific prompts together with image tokens into CLIP is illustrated in Fig. \ref{fig2}. Let $P_c$ represent domain-shared prompts, $P_k$ represent task-specific prompts from pre-trained prompt network, $e_0$ is the embedding of an additional learnable classification token ([CLS]), and $\mathcal{E}$ is a sequence of embedding image patches. The input to CLIP can be written as
\begin{equation}
\label{e1}
\begin{split}
    x = [ e_0, f_1(P_c), f_2(P_k), \mathcal{E} ],
\end{split}
\end{equation}
where $[\cdot, \cdot]$ represents the concatenation operation, and both $f_1$ and $f_2$ represent mapping functions implemented through MLP are used to embed prompt tokens.

In the downstream tasks, as shown in Fig.\ref{fig2}, the entire CLIP network is frozen, and the prompt parameters (including learnable domain-shared prompt tokens $P_c$, and its following MLP parameters $\theta_m$, the projection parameter $\theta_p$ of task-specific prompts $P_k$, and the head parameters $\theta_h$) are optimized, which are denoted as $\Theta = (P_c, \theta_m, \theta_p, \theta_h)$ in the following for clarity. Then, the prompt parameters $\Theta$ can be solved by
\begin{equation}
\label{e3}
    \Theta^{*} = \arg \min_{\Theta}{\mathcal{L}(D_\mathcal{S},D_\mathcal{T})},
\end{equation}
where $\mathcal{L}$ is the meta-prompt loss function, $D_\mathcal{S}$ and $D_\mathcal{T}$ represent data sets for the source and target domains, respectively. %The meta-trained parameters are passed into the meta-test phase for use. When testing new tasks, the learned prompts can be learned with the input image to be a new input for a more accurate classification. 

In the following, our primary task is to learn the prompt parameters $\Theta$ in an efficient way. In summary, we design a meta prompt learning framework in the meta-training phase, which will be introduced in Section \ref{empl_base}. Then, in the meta-test phase, the learned prompt parameters along with the task-specific data are utilized as a new input for accurate classification in new tasks.

\subsection{Proposed Meta Prompt Learning Framework}\label{empl_base}
We now develop a meta prompt learning framework, E$^2$MPL, to learn domain-shared prompts and task-specific prompts from meta-tasks such that the model can adapt to new tasks.
Specifically, the learning process can be formulated as a bilevel optimization problem, where the inner objective is conducted on the support set to learn a classifier and a domain adapter that handle the current task, while the outer objective is conducted on the query set to update the prompt parameters for storing the corresponding meta-knowledge.

\textbf{(1) For the inner objective:} 

%For each meta task, we fix the feature embedding parameter $\theta_e$ to obtain the feature embeddings $Z_{\mathcal{S},i}^\text{sup}$ and $Z_{\mathcal{ST},i}^\text{sup}$ previously defined. Then, the base learners $\mathcal{A}$ and $\mathcal{B}$ are trained to obtain the optimal parameters $\theta_{\omega}^{*}$ and $\theta_{d}^{*}$, which can be written by:

% For each meta task i, we fix the CLIP parameter $\theta_e$, and meanwhile train the two base learners $\mathcal{A}$ and $\mathcal{B}$ to obtain their optimal parameters $\theta_{\omega}^{*}$ and $\theta_\mathcal{T}^{*}$. 
For each meta task $i$, with the prompt parameters fixed, we design the two base learners ($\mathcal{A}$ for classification and $\mathcal{B}$ for domain alignment) to calculate their optimal parameters $\theta_{\omega}^{*}$ and $\theta_\mathcal{T}^{*}$, respectively. 
% The learner A is a classifier built by labeled source domain support set $D_i^{sup}$ in internal calculation, and the learner B is a domain adapter obtained from source domain support set $D_i^{\text{sup}}$ and target domain query set $D_i^{\text{que}}$. 
% Both the classifier and the domain adapter are derived from the closed-form solution computed by the ridge regression formula, which will be described in Sections \ref{rr_base} and \ref{ts_base}. By calculating the closed-form solution, ridge regression can quickly calculate the optimal classifier parameters. 
This efficiency makes it possible to quickly obtain classifiers in few-shot learning scenarios without complex iterative optimization processes.
Let $Z_{\mathcal{S},i}^\text{sup}$ and $Z_{\mathcal{T},i}^\text{que}$ represent the feature embeddings that are used for classification and domain adapter. Then, the inner objective functions can be written by:
\begin{equation}%\small
\label{e4}
\begin{split}
\theta_{\omega}^{*} =& \arg \min_{\theta_{\omega}} \mathcal{A}(D_i^{\text{sup}}; \theta_{\omega}) \\
=& \arg \min_{\theta_{\omega}} \mathcal{L}_c^{\text{base}}(Z_{\mathcal{S},i}^\text{sup}, Y_{\mathcal{S},i}^\text{sup}; \theta_{\omega}), \\
% & + \mathcal{L}_F^{\text{base}}(Z_{\mathcal{S},i}^\text{sup}, Y_{\mathcal{S},i}^\text{sup}; \theta_{\omega}),\\
\end{split}
\end{equation}
\begin{equation}%\small
\label{e5}
\begin{split}
% \theta_{d}^{*} =& \arg \min_{\theta_{d}}\mathcal{B}(D_i^{\text{sup}},Y_{d_i}^{\text{sup}}; \theta_d)\\ =& \arg \min_{\theta_{d}} \mathcal{L}_d^{\text{base}}(Z_{\mathcal{ST},i}^\text{sup},Y_{d_i}^{\text{sup}}; \theta_{d}),\\
\theta_\mathcal{T}^{*} =& \arg \min_{\theta_\mathcal{T}}\mathcal{B}(D_i^{\text{sup}},D_i^{\text{que}}; \theta_\mathcal{T})\\ =& \arg \min_{\theta_\mathcal{T}} \mathcal{L}_{d}^{\text{base}}(Z_{\mathcal{S},i}^\text{sup},Z_{\mathcal{T},i}^\text{que}; \theta_\mathcal{T}). \\
% & + \mathcal{L}_F^{\text{base}}(Z_{\mathcal{S},i}^\text{sup}, Y_{\mathcal{S},i}^\text{sup}; \theta_{\mathcal{T}}),\\
\end{split}
\end{equation}
% where $\mathcal{L}_c^{\text{base}}(\cdot)$ represent the loss functions of the base learner, and will be introduced in Sections \ref{fc_base} and \ref{rr_base}. In Eq. (\ref{e2}), $Y_{d_i}^{\text{sup}}$ represents the matrix of domain labels of support samples where source and target domains are set as $0$ and $1$, respectively.
Here, $\mathcal{L}_c^{\text{base}}(\cdot)$ and $\mathcal{L}_{d}^{\text{base}}(\cdot)$
%$\mathcal{L}_{F}^{\text{base}}$ 
represent the loss functions of classification and domain alignment, which will be described in Sections \ref{rr_base} and \ref{ts_base}, respectively. %In this phase, unlike the traditional meta-prompt learning, we use the calculation of the closed-form solution instead of multiple optimization, which greatly reduces the training time.%, and will be introduced in Sections \ref{rr_base} and \ref{ts_base}. % In Eq. (\ref{e2}).
% and $\mathcal{L}_d^{\text{base}}(\cdot)$ 

\textbf{(2) For the outer objective:} 
% After obtaining the optimal $\theta_{\omega}^{*}$ and $P_n^{*}$ for each meta task, we update the prompts $P_c$ and other learnable parameters, by minimizing the classification loss $\mathcal{L}_c^{\text{meta}}$ and adversarial loss $\mathcal{L}_{d}^{\text{meta}}$ (against the domain discriminator) on the query set. Formally, the outer objective function can be written by:

% After obtaining the optimal $\theta_{\omega}^{*}$ and $\theta_{\mathcal{T}}^{*}$ for each meta task, we update the prompt module and other learnable parameters, by minimizing the classification loss $\mathcal{L}_c^{\text{meta}}$, entropy minimization loss $\mathcal{L}_{EN}^{\text{meta}}$ and classes discrimination loss $\mathcal{L}_{F}^{\text{meta}}$ on the support set and query set. Formally, the outer objective function can be written by:
After obtaining the (inner) optimal $\theta_{\omega}^{*}$ and $\theta_{\mathcal{T}}^{*}$ for each meta task $i$, we update the prompt parameters $\Theta$ by minimizing the meta prompt loss, which consists of the classification loss $\mathcal{L}^{\text{meta}}_c$, the entropy minimization loss $\mathcal{L}^{\text{meta}}_{d}$, and the class discrimination loss $\mathcal{L}^{\text{meta}}_{f}$. % \emph{i.e.,} 
% \begin{equation}
% \label{e2}
%     \mathcal{L} = \mathcal{L}_c^{\text{meta}} + \lambda_d \mathcal{L}_{d}^{\text{meta}} + \lambda_f \mathcal{L}_f^{\text{meta}},
% \end{equation}
% where $\lambda_d$ and $\lambda_f$ are hyperparameters that regulate the balance between the losses. 
Specifically, 
%We first encode the query set and the learnable prompts through the pre-trained CLIP and head to obtain feature embeddings. Then, the feature embeddings obtain logits through classifier A.To guide the model training, 
the classification loss $\mathcal{L}^{\text{meta}}_c$ is calculated by using $\theta_{\omega}^{*}$ to classify the query set $D_i^{\text{que}}$ from the source domain. 
Furthermore, due to the domain gap between the target and source domain data, we use $\theta_{\mathcal{T}}^{*}$ to project the target domain data into the feature space close to the source domain data. Then, the entropy minimization loss $\mathcal{L}^{\text{meta}}_{d}$ is calculated for the projected target domain data. 

In addition, we also design a class discrimination loss $\mathcal{L}^{\text{meta}}_{f}$, which minimizes the intra-class scatter matrix $S_w$ and maximizes the inter-class scatter matrix $S_b$ to bring the same-class samples closer together while pushing those of different classes farther apart, thereby improving the classification performance, which can be written by:  
\begin{equation}
\label{e_new}
\mathcal{L}_f^{\text{meta}}=S_w-\lambda_sS_b,
\end{equation}
where $\lambda_s$ is a learnable hyperparameter. Since the target domain data are unlabeled, we use the cosine similarity between the target domain query samples and the source domain support samples to define the "same-class" samples. If the similarity is greater than a certain threshold, it is considered to be of the same class; otherwise, it is considered different. The threshold is automatically learned by the model (usually in the range of 0.65-0.8).

Formally, the losses will be back-propagated to update the prompt parameters $\Theta$. Thus, the outer objective function can be written by:
\begin{equation}%\small
\label{e6}
\begin{split}
% &\min_{\theta_e}\ \mathcal{L}_c^{\text{meta}}(D_i^{\text{que}};\theta_e,\theta_{\omega}^{*})+\lambda\mathcal{L}_d^{\text{meta}}(D_i^{\text{que}},Y_{d_i}^{\text{que}};\theta_e,\theta_{d}^{*}),\\
% &\ \text{with} \; \theta_{\omega}^{*}\!=\!\mathcal{A}(D_i^{\text{sup}}; \theta_e) \ \ \text{and}\ \ \theta_{d}^{*}\!=\!\mathcal{B}(D_i^{\text{sup}}, Y_{d_i}^{\text{sup}}; \theta_e),
\min_{\Theta}&\ \mathcal{L}_c^{\text{meta}}(D_i^{\text{que}};\Theta,\theta_{\omega}^{*})+\lambda_d\mathcal{L}_{d}^{\text{meta}}(D_i^{\text{que}};\Theta,\theta_{\mathcal{T}}^{*}) \\
& + \lambda_f\mathcal{L}_{f}^{\text{meta}}(D_i^{\text{que}}; \Theta,\theta_{\mathcal{\omega}}^{*}, \theta_{\mathcal{T}}^{*}), \\
&\ \text{with} \; \theta_{\omega}^{*}\!=\!\mathcal{A}(D_i^{\text{sup}}; \Theta) \\   
&\ \text{and}\ \ \theta_{\mathcal{T}}^{*}\!=\!\mathcal{B}(D_i^{\text{sup}}, D_i^{\text{que}}; \Theta),
\end{split}
\end{equation}
where $\lambda_d$ and $\lambda_f$ are hyperparameters that regulate the balance between the losses.
%where $\lambda$ is the parameter to balance the two loss terms. %\re{and will be analyzed in the supplementary material}. 
%Opposite to $Y_{d_i}^{\text{sup}}$, $Y_{d_i}^{\text{que}}$ represents the matrix of domain labels of the query samples where source and target domains are set as $1$ and $0$, respectively. 
% In this phase, we first leverage $\theta_{\omega}^*$ and $\theta_{\mathcal{T}}^*$ to predict the classification labels and domain labels of samples in the query set, respectively. Then, we use the cross-entropy loss to calculate $\mathcal{L}_c^{\text{meta}}$ and $\mathcal{L}_d^{\text{meta}}$ between the predicted and true labels. Finally, the two losses are back-propagated to update the feature embedding parameter $\theta_e$ by using gradient descent. The two above phases are alternately conducted on all meta tasks in an episodic training way. 
% In this phase, we first leverage $\theta_{\omega}^*$ and $\theta_{\mathcal{T}}^*$ to predict the classification labels and realize domain alignment in the query set, respectively. Then, we use the cross-entropy loss to calculate $\mathcal{L}_c^{\text{meta}}$ between the predicted and true labels, the entropy minimization loss to calculate $\mathcal{L}_{EN}^{\text{meta}}$ on projected query set for target domain and classes discrimination loss to calculate $ \mathcal{L}_{F}^{\text{meta}}$ on support set and its labels during training and testing. Finally, the three losses are back-propagated to update the paremeters $\Theta$.
The optimal task-specific parameters $\theta_{\omega}^*$ and $\theta_{\mathcal{T}}^*$ obtained in the inner objective are used for classification and domain alignment in the query set, respectively.

Accordingly, the prompt parameters $\Theta$ in the outer objective will be iteratively updated in multiple meta-tasks to acquire meta-knowledge for rapid adaptation to new tasks.

\subsection{Ridge Regression for Classification}\label{rr_base}
%For computational efficiency, we leverage ridge regression to build base learners.
%$\mathcal{A}$ (for classification) and $\mathcal{B}$ (for domain discrimination). 
%It has a globally optimal closed-form solution. 
Traditional meta-learning often involves multiple optimizations in the inner loop to update parameters, which is time-consuming and poses a significant challenge for quickly and effectively adapting to new classes. Ridge regression, with its closed-form solution, regularization, differentiability, and computational efficiency, becomes a good choice for efficiently adapting to new classes in few-shot learning. For computational efficiency and inspired by recent work\cite{DBLP:conf/iclr/BertF19}, we use ridge regression to construct the base learners for each meta task. The optimal parameters $\theta_{\omega}^{*}$ can be directly calculated as % and $\theta_{d}^{*}$ 
\begin{equation}
%\small
\label{e7}
\begin{split}
\theta_{\omega}^{*} =&\arg\min _{\theta_{\omega}} \left\|Z_{\mathcal{S},i}^{\text{sup}} \theta_{\omega}-Y_{\mathcal{S},i}^{\text{sup}}\right\|^{2}+\gamma_{\omega}\left\|\theta_{\omega}\right\|^{2}\\
=&\left({Z_{\mathcal{S},i}^{\text{sup}}}^\top Z_{\mathcal{S},i}^{\text{sup}}+\gamma_{\omega} I\right)^{-1} {Z_{\mathcal{S},i}^{\text{sup}}}^\top Y_{\mathcal{S},i}^{\text{sup}},\\
\end{split} 
\end{equation}
% \begin{displaymath}\small\label{e5}
% % \setlength\abovedisplayskip{2pt}
% % \setlength\belowdisplayskip{2pt}
% \begin{split}
% \theta_{d}^{*}=&\mathcal{B}(D_i^{\text{sup}}, Y_{d_i}^{\text{sup}}; \theta_e)=\arg\min _{\theta_{d}} \left\|Z_{\mathcal{ST},i}^{\text{sup}} \theta_{d}-Y_{d_i}^{\text{sup}}\right\|^{2}+\gamma_d\left\|\theta_{d}\right\|^{2}\\
% =&\left({Z_{\mathcal{ST},i}^{\text{sup}}}^\top Z_{\mathcal{ST},i}^{\text{sup}}+\gamma_d I\right)^{-1} {Z_{\mathcal{ST},i}^{\text{sup}}}^\top Y_{d_i}^{\text{sup}},\\
% \end{split}
% \end{displaymath}
where $\gamma_{\omega}$ is a learnable regularization parameter. The above matrix to inverse are in $m\times m$ ($m$ is the dimensionality of feature embeddings).  % , \gamma_d
%Nevertheless, since a deep feature embedding model usually generates high-dimensional features, the matrix inversion could be computationally expensive.
% By the Woodbury's identity \cite{Petersen2008matrix}, the above equations can be reformulated as
To improve computational efficiency and reduce memory requirements, we use the Woodbury identity \cite{Petersen2008matrix} to reformulate the above equation as
\begin{equation}%\small
\label{e8}
\theta_{\omega}^{*} = {Z_{\mathcal{S},i}^{\text{sup}}}^\top\left(Z_{\mathcal{S},i}^{\text{sup}}{Z_{\mathcal{S},i}^{\text{sup}}}^\top+\gamma_{\omega} I\right)^{-1} Y_{\mathcal{S},i}^{\text{sup}},
\end{equation}
% \begin{equation}\label{e7}
% % \setlength\abovedisplayskip{2pt}
% % \setlength\belowdisplayskip{3pt}
% \theta_{d}^{*} = {Z_{\mathcal{ST},i}^{\text{sup}}}^\top\left(Z_{\mathcal{ST},i}^{\text{sup}}{Z_{\mathcal{ST},i}^{\text{sup}}}^\top+\gamma_d I\right)^{-1}Y_{d_i}^{\text{sup}},
% \end{equation}
where the matrix to inverse reduces to $n\times n$ ($n$ is the number of support samples in a task). Because $n$ is much less than ${m}$ in the FS-UDA setting, $\theta_{\omega}^{*}$ can be efficiently calculated. %In this way, our E$^2$MPL model can be efficiently trained and adapted to new tasks. 

\subsection{Cross-domain Projection for Domain Adaptation}\label{ts_base}
% It is evident that the data in the meta-task has significant domain gaps. Except for the classification task same as in Section \ref{rr_base}, domain alignment is required in each meta task. We aim to learn a direct transformation projection from target domain to source domain, aligning the two domains. 
As the query set and the support set come from different domains, we design a domain adapter to align the domains in each meta task. Typically, domain adapters introduce a learnable transformation layer that effectively adjusts the feature distribution between the two domains \cite{da}. Since the target domain data are not labeled, direct cross-domain distribution alignment can easily cause class misalignment (different classes are pulled together). Thus, we design a domain projection parameter $\theta_\mathcal{T}$ to directly transform the target domain data into a feature space close to the source domain data. Specifically, we first measure the similarity of samples between the target domain and the source domain, and then further pull the target domain samples and similar source domain samples closer. 

\begin{algorithm}[t]
%\scriptsize
\caption{Proposed E$^2$MPL}
\label{alg:algorithm}
\hspace*{0.02in}{\textbf{Training Input}}: An auxiliary dataset %including labeled source domain data $\mathcal{S}_{\text{aux}}$ and 
including labeled $\mathcal{S}_{\text{aux}}$ and unlabeled  $\mathcal{T}_{\text{aux}}$, and learning rate $\beta$\\
\hspace*{0.02in}{\textbf{Training Output}}: prompt parameters $\Theta$

\begin{algorithmic}[1]
%\STATE Randomly initialize $\theta_e$.
\WHILE{not converged}
\STATE Sample a meta task consisting of $D_i^\text{sup}$ and $D_i^\text{que}$ from the auxiliary dataset.\\
\STATE\textcolor{cyan}{\COMMENT{\texttt{obtain the feature embeddings}}}
$Z_{\mathcal{S},i}^\text{sup}=f_e(X_{\mathcal{S},i}^{\text{sup}};\Theta), Z_{\mathcal{T}, i}^{\text{que}} = f_e(X_{\mathcal{T},i}^{\text{que}};\Theta) $%, Z_{\mathcal{ST},i}^\text{sup} = f_e(X_{\mathcal{S},i}^{\text{sup}},X_{\mathcal{T},i}^{\text{sup}};P_c)$
\STATE\textcolor{cyan}{\COMMENT{\texttt{optimize classification parameter}}}
$\theta_\omega^*\leftarrow$ closed-form solution of Eqn. (\ref{e8}). 
% \STATE\textcolor{cyan}{\COMMENT{\texttt{optimize discrimination parameter}}}\\
% $\theta_d^*\leftarrow$ closed-form solution of Eqn. (\ref{e7}).
\STATE\textcolor{cyan}{\COMMENT{\texttt{optimize domain projection parameter}}}
$\theta_\mathcal{T}^*\leftarrow$ closed-form solution of Eqn. (\ref{e11}).
\STATE Predict the labels $\hat{Y}_{\mathcal{S},{\text{i}}}^{\text{que}}$ and $\hat{Y}_{\mathcal{T},{\text{i}}}^{\text{que}}$ of the query set by applying $\theta_\omega^*$ and $\theta_\mathcal{T}^*$
% $Z_{\mathcal{S}, i}^{\text{que}}P_n^*\theta_\omega^*$ and $\hat{Y}_{d_i}^{\text{que}}$  and $\theta_d^*$
\STATE Solve Eqn. (\ref{e6}) by calculating its gradient w.r.t. $\Theta$ %  and $\mathcal{L}_{d}^{\text{meta}}$
% $\nabla_{\Theta}(\mathcal{L}_{c}^{\text{meta}}+\lambda_d\mathcal{L}_{d}^{\text{meta}}+\lambda_f\mathcal{L}_{f}^{\text{meta}})$.
\STATE\textcolor{cyan}{\COMMENT{\texttt{update feature embedding parameter}}}
$\Theta \leftarrow \Theta - \beta\nabla_{\Theta}(\mathcal{L}^{\text{meta}}_c+\lambda_d\mathcal{L}^{\text{meta}}_{d}+\lambda_f\mathcal{L}_{f}^{\text{meta}})$.
\ENDWHILE
\end{algorithmic}
% \textcolor{cyan}{\COMMENT{\texttt{Test on a new task}}}
\hspace*{0.02in}{\textbf{Testing Input}}: $D_{\mathcal{S}}^{\text{sup}}$ and $D_{\mathcal{T}}^{\text{que}}$ in a new task and learned $\Theta$\\
\hspace*{0.02in}{\textbf{Testing Output}}: Prediction $\hat{Y}_{\mathcal{T},{\text{new}}}^{\text{que}}$ for this new task

\begin{algorithmic}[1]
\STATE Calculate  $Z_{\mathcal{S},{\text{new}}}^\text{sup}=f_e(X_{\mathcal{S},{\text{new}}}^{\text{sup}};\Theta)$ and\\ 
$\theta_\omega^{\text{new}} \leftarrow {Z_{\mathcal{S},{\text{new}}}^{\text{sup}}}^\top\left(Z_{\mathcal{S},{\text{new}}}^{\text{sup}}{Z_{\mathcal{S},{\text{new}}}^{\text{sup}}}^\top+\gamma_{\omega} I\right)^{-1} Y_{\mathcal{S},{\text{new}}}^{\text{sup}}$.
\STATE Calculate $Z_{\mathcal{T},{\text{new}}}^{\text{que}}=f_e(X_{\mathcal{T},{\text{new}}}^{\text{que}};\Theta)$ and\\
$\theta_\mathcal{T}^{\text{new}} \leftarrow {Z_{\mathcal{T},{\text{new}}}^{\text{que}}}^\top\left(D^AZ_{\mathcal{T},{\text{new}}}^{\text{que}}{Z_{\mathcal{T},{\text{new}}}^{\text{que}}}^\top+\gamma_p I\right)^{-1} AZ_{\mathcal{S},{\text{new}}}^{\text{sup}}$.
%$\theta_p \leftarrow {Z_{\mathcal{T},{\text{new}}}^{\text{que}}}^\top\left(D^AZ_{\mathcal{T},{\text{new}}}^{\text{que}}{Z_{\mathcal{T},{\text{new}}}^{\text{que}}}^\top+\gamma_p I\right)^{-1} AZ_{\mathcal{S},{\text{new}}}^{\text{sup}}$.}
\STATE Predict the labels $\hat{Y}_{\mathcal{T},{\text{new}}}^{\text{que}}$ by using $Z_{\mathcal{T},{\text{new}}}^{\text{que}}\theta_\mathcal{T}^{\text{new}}\theta_\omega^{\text{new}}$.
\end{algorithmic}
\end{algorithm}

Let $A$ stand for the similarity matrix of samples in the query set from target domain to the support set from source domain. Then, $D^A$ represents the diagonal matrix of all row-summing values of $A$. 
Formally, in each meta task, the optimization problem w.r.t. $\theta_\mathcal{T}$ can be written by:
\begin{equation}\label{e9}
% \small
\begin{split}
\min_{\theta_\mathcal{T}}&\sum_{ik}A_{ik}||Z_{\mathcal{T},i}^{\text{que}}\theta_\mathcal{T}-Z_{\mathcal{S},k}^{\text{sup}}||^2+\gamma_p||\theta_\mathcal{T}||^2\\
=tr&(\theta_\mathcal{T}^{\top}Z_{\mathcal{T}}^{\text{sup}\top}D^AZ_{\mathcal{T}}^{\text{sup}}\theta_\mathcal{T})-2tr(\theta_\mathcal{T}^{\top}Z_{\mathcal{T}}^{\text{sup}\top}AZ_{\mathcal{S}}^{\text{sup}})+\gamma_p||\theta_\mathcal{T}||^2,\\
\end{split}
\end{equation}
% Where, $\gamma_p$is the regularization parameter.
where $Z_{\mathcal{T},i}^{\text{que}}\theta_\mathcal{T}$ represents the projected feature of the $i$-th target domain sample $Z_{\mathcal{T},i}^{\text{que}}$, and $A_{ik}$ is its similarity to the $k$-th source domain sample $Z_{\mathcal{S},k}^{\text{sup}}$. We use the regularizer $||\theta_\mathcal{T}||^2$ to prevent overfitting, and $\gamma_p$ is a learnable regularization parameter. Then, by setting the partial derivation of Eq. (\ref{e9}) w.r.t. $\theta_\mathcal{T}$ as zero, we obtain 
\[
\begin{split}
2Z_{\mathcal{T}}^{\text{que}\top}D^AZ_{\mathcal{T}}^{\text{que}}\theta_\mathcal{T}-2Z_{\mathcal{T}}^{\text{que}\top}AZ_{\mathcal{S}}^{\text{sup}}+2\gamma_p\theta_\mathcal{T}=0,\\
(Z_{\mathcal{T}}^{\text{que}\top}D^AZ_{\mathcal{T}}^{\text{que}}+\gamma_p I_c)\theta_\mathcal{T}=Z_{\mathcal{T}}^{\text{que}\top}AZ_{\mathcal{S}}^{\text{sup}}.
\end{split}
\]
Thus, $\theta_\mathcal{T}$ has a closed-form solution, which can be written as
\begin{equation}\label{e10}
\theta_\mathcal{T}^{*}=(Z_{\mathcal{T}}^{\text{que}\top}D^AZ_{\mathcal{T}}^{\text{que}}
+\gamma_p I_c)^{\text{-1}}Z_{\mathcal{T}}^{\text{que}\top}AZ_{\mathcal{S}}^{\text{sup}}.
\end{equation}
For computation efficiency, we leverage Woodbury identity to update the above solution as:
\begin{equation}\label{e11}
\theta_\mathcal{T}^{*}=Z_{\mathcal{T}}^{\text{que}\top}(D^AZ_{\mathcal{T}}^{\text{que}}Z_{\mathcal{T}}^{\text{que}\top}+\gamma_p I_n)^{\text{-1}}AZ_{\mathcal{S}}^{\text{sup}},
\end{equation}
where the matrix $A$ is calculated by
$A_{ik}=e^{-{||Z^{\text{que}}_{\mathcal{T},i}-Z^{\text{sup}}_{\mathcal{S},k}||}^2}$. To avoid the value explosion caused by matrix multiplication, we alternately normalize the matrix $A$ row by row and column by column until convergence. Note that alternating normalization of rows and columns not only improves the stability and efficiency of the calculation, but also reduces numerical instability when dealing with large-scale data.

In summary, the domain projection parameter $\theta_\mathcal{T}^{*}$ can be calculated in one step to narrow the domain gap. Thus, labeled data from the source domain can be leveraged to enhance the classification of unlabeled data in the target domain. 

In this way, thank to closed-form solutions of $\theta_\omega^{*}$ and $\theta_\mathcal{T}^{*}$, the proposed E$^2$MPL model can be efficiently trained among meta-tasks and quickly adapted to new tasks.
The meta-train and meta-test process is summarized in Algorithm~\ref{alg:algorithm}. During the meta-train, given pretrained encoder and the sampled meta-task $i$ from an auxiliary dataset, the algorithm first calculates the optimal parameters $\theta_\omega^*$ and $\theta_\mathcal{T}^*$ for classification and domain alignment via the closed-form solutions and then validates them in the query set. Afterwards, the resulted meta prompt loss is used to update the prompt parameters $\Theta$ by gradient descent. 
During the meta-test on a new task, with the learned prompts, the algorithm calculates the new $\theta_\omega^{\text{new}}$ and $\theta_\mathcal{T}^{\text{new}}$ to test the classification in the target domain.

\section{Experiments}
%In this section, we introduce the datasets, implementation details and setup. To validate the efficacy of our MELDA for FS-UDA, we compare it with eleven baselines. Moreover, we experiment the effect of different domain discriminators and ablation study. \textbf{In addition, parameter analysis and more results can be found in the supplementary material.}
In this section, we demonstrate the effectiveness of our proposed E$^2$MPL. We first introduce datasets, implementation details and meta-task setting. Next, we compare the proposed method with other related methods to evaluate the ability of adaptation to target domain, and task generalization to unseen categories. Then, we provide the ablation studies to explore the contribution of each component in E$^2$MPL. We also design a comparison of training time and provide analysis to show the efficacy of the proposed method.
\subsection{Datasets and Implementation}
We experiment on the benchmark dataset 
% \emph{miniImageNet} \cite{DBLP:conf/iclr/RaviL17}, \emph{Office-Home} \cite{DBLP:journals/corr/VenkateswaraECP17} and 
\emph{DomainNet} \cite{domainnet}, which 
% \emph{miniImageNet} is popularly used in few-shot learning, \emph{Office-Home} is widely used for domain adaptation and \emph{DomainNet} 
has usually been used for FS-UDA methods. It was released in 2019 for the research of multi-source domain adaptation \cite{domainnet}, and consists of six distinct domains: \emph{quickdraw, clipart, real, sketch, painting, infograph}, each containing a diverse array of images categorized into 345 object classes. In our experiments, we use the first five domains and first select any two of the five domains as the source and target domains, respectively, thus forming 20 different source-to-target domain adaptation tasks. Then, we discard the 40 categories that contain less than 20 images for the two domains. Finally, we randomly split the images of the remaining 308 categories into images of 217 categories, 43 categories, and 48 categories, and adjusted the size of these images to 224 $\times$ 224 pixels for meta-training (forming the auxiliary dataset), model validation and testing new tasks, respectively. %`*' indicates that we modify \emph{DomainNet} to fit our setting (see the paragraph of datasets for details).

\textbf{Implementation details.}
Our experiments are carried out with a single NVIDIA GTX 3090 GPU and PyTorch. The network architecture of our E$^2$MPL consists of two components: the backbone and the prompt module. We utilize two different backbones: a pretrained ViT-B/32 (CLIP) with prompts used in \cite{CLIP} and a pretrained 12-layer ResNet network (ResNet-12) used in \cite{leeMetaCVPR2019}. For pretraining of the CLIP, we only used the fixed visual encoder ViT-B/32, and to speed up training, we used the head layer to reduce the data feature dimension from 512 to 128.
For pretraining of the ResNet-12, we use ResNet-12 as a feature extractor and full connected layer as a classifier for pretraining on \emph{miniImageNet}. In addition, since the feature extraction capability of ResNet-12 is not as powerful as that of CLIP, we introduce a data augmentation method \cite{OSA}\cite{DOMFN}\cite{PEMW} when the backbone is ResNet-12. The prompt module consists mainly of domain-shared prompts, task-specific prompts, and a head layer. Through experiments, we found that the best experimental results were achieved when the domain-shared prompt was composed of four prompts. Therefore, the size of the domain-shared prompt is 4 $\times$ 768 (768 is the dimension of the image patch). For task-specific prompts, we also use ResNet-12 to pre-train on \emph{miniImageNet} to obtain prior knowledge. Afterwards, we train E$^2$MPL for 10 epochs and perform episodic training on 1000 episodes for each epoch. In each episode, a meta-task is randomly constructed from the dataset to train our meta-prompt model. For model training, we use the Adam solver with the related gradient weights of $0.5$ and $0.999$, and the learning rate is empirically set as $0.005$. Both $\lambda_d$ and $\lambda_f$ in Eqn. (\ref{e6}) are set to $0.01$ for \emph{DomainNet}. The $\gamma_\omega$ and $\gamma_p$ in Eqns. (\ref{e8}) and (\ref{e11}) are automatically updated along with the meta prompt learner through the outer optimization. During the meta-test, we randomly select 3600 tasks to calculate the averaged top-1 accuracy as the evaluation criterion.

\textbf{To realize the $5$-way $K$-shot UDA setting}, in each meta task, we first randomly samples $5*(K+15)$ labeled samples from the source domain and $75$ unlabeled samples from the target domain, which are derived from five categories randomly selected equally. Then, the $(K+15)$ source domain samples in each category are randomly partitioned into $K$ for the support set and $15$ for the query set, while the $75$ target domain samples are used for the query set. For each meta-test task, its support and query sets have the same set of five new categories. The support set contains $5K$ labeled source domain samples for classifier update, while the query set contains $75$ unlabeled target domain samples for performance evaluation.

\begin{table*}%[t]
\caption{Comparison of our E$^2$MPL and the baseline methods for the FS-UDA setting. The classification accuracy (\%) of target domain samples are averaged over 3600 new tasks under the $5$-way, $K$-shot setting. %The backbone we used is the CLIP (ViT-B/32) and completed the experiments. 
The best performances are shown in bold.}
\label{tab:table222}
\centering
%\scriptsize
\renewcommand{\arraystretch}{1.2}
\setlength{\tabcolsep}{6pt}
% \vspace{-0.4cm}
\resizebox{\linewidth}{!}{
\begin{tabular}{c|cccccccccc|c}
\hline\hline
\multicolumn{11}{c}{\textbf{5-way, 1-shot, backbone: CLIP (ViT-B/32)}} \\
\hline \multirow{2}{*}{\textbf{Methods}}
    &\textbf{skt$\leftrightarrow$rel} &\textbf{skt$\leftrightarrow$qdr} &\textbf{skt$\leftrightarrow$pnt} &\textbf{skt$\leftrightarrow$cli} &\textbf{rel$\leftrightarrow$qdr} &\textbf{rel$\leftrightarrow$pnt} &\textbf{rel$\leftrightarrow$cli} &\textbf{qdr$\leftrightarrow$pnt} &\textbf{qdr$\leftrightarrow$cli} &\textbf{pnt$\leftrightarrow$cli} & \textbf{avg}\\
     &$\rightarrow$ / $\leftarrow$&$\rightarrow$ / $\leftarrow$&$\rightarrow$ / $\leftarrow$&$\rightarrow$ / $\leftarrow$&$\rightarrow$ / $\leftarrow$&$\rightarrow$ / $\leftarrow$&$\rightarrow$ / $\leftarrow$&$\rightarrow$ / $\leftarrow$&$\rightarrow$ / $\leftarrow$&$\rightarrow$ / $\leftarrow$ & -- \\\hline
\textbf{MCD} \cite{DBLP:conf/cvpr/SaitoWUH18}
     &71.64/63.03 &39.71/38.02 &63.24/60.26 &59.52/65.28 &34.97/59.86 &67.19/68.65 &65.45/77.64 &43.05/31.39 &57.24/37.41 &59.94/69.61 &56.66\\
\textbf{ADDA} \cite{DBLP:conf/cvpr/TzengHSD17}
     &73.97/64.36 &42.17/39.73 &61.85/61.26 &60.03/67.56 &35.44/63.72 &69.09/69.93 &65.14/76.51 &43.61/34.68 &58.40/38.13 &66.92/72.18 &58.23\\
\textbf{DWT}  \cite{roy2019unsupervised}
     &73.21/63.86 &43.59/42.45 &63.94/65.57 &61.06/68.36 &35.49/62.81 &69.68/71.32 &67.83/79.15 &48.02/35.19 &58.87/39.47 &61.38/74.79 &59.30\\
\hline
\textbf{MAML} \cite{DBLP:conf/icml/FinnAL17}
    &77.88/76.93 &51.34/50.25 &69.67/67.48 &75.89/76.29 &48.39/51.69 &75.86/74.82 &79.74/80.25 &44.04/43.48 &53.31/51.98 &68.86/68.95 &64.36\\
\textbf{R2D2} \cite{DBLP:conf/iclr/BertF19}
    &72.10/79.19 &56.89/58.10 &71.09/72.69 &80.32/79.95 &51.84/55.06 &76.58/79.17 &82.67/84.29 &48.57/45.70 &60.63/55.79 &74.79/74.12 &67.98\\
\textbf{Meta-ticket}\cite{DBLP:conf/neurips/ChijYA22}
    &79.19/78.98 &56.23/53.94 &71.68/69.37	 &78.26/78.67 &51.48/52.89 &75.91/75.05	 &80.72/81.22 &45.63/43.51 &56.97/57.49	 &70.75/68.42 &66.32\\
\hline
\textbf{baseline} \cite{chen2019closer}
    &65.16/59.05 &40.59/39.30 &56.15/55.40 &62.90/65.22 &37.55/39.47 &55.44/62.23 &63.70/72.52 &33.35/34.38 &40.90/40.44 &57.08/59.52 &52.02\\
\textbf{baseline++} \cite{chen2019closer}
    &73.00/74.00 &45.49/53.68 &65.32/64.32 &70.64/71.44 &43.98/50.86 &72.30/69.29 &76.70/77.81 &42.33/36.79 &53.86/46.01 &65.18/65.50 &60.93\\
\textbf{ProtoNet} \cite{DBLP:journals/corr/SnellSZ17}
    &77.40/75.97 &51.60/50.87 &69.42/70.04 &76.83/77.23 &45.29/49.60 &73.06/75.18 &78.85/80.93 &40.51/44.61 &52.88/51.51 &71.53/69.99 &64.17\\
\textbf{DN4} \cite{li2019revisiting}
    &72.97/65.17 &39.39/57.99 &63.68/61.31 &62.45/69.55 &34.25/61.19 &68.70/70.04 &66.79/78.79 &52.11/34.58 &59.93/40.35 &60.94/77.00 &59.86\\
\textbf{ADM} \cite{li2020asymmetric}
    &73.25/65.94 &42.81/59.42 &64.93/64.57	 &64.39/72.67 &45.82/60.96 &70.86/72.63	 &67.04/79.91 &48.77/35.14 &59.21/43.87	 &62.90/76.56 &61.58\\
\textbf{FEAT} \cite{FEAT}
    &74.75/67.74 &44.59/58.60 &66.19/62.80	  &69.90/72.51 &51.18/60.71 &69.69/71.06	 &77.12/79.59 &47.37/43.73 &60.67/52.41	 &64.39/70.01 &63.25\\
\textbf{DeepEMD} \cite{DeepEMD}
    &73.54/67.02 &43.15/58.48 &65.73/63.62	 &68.09/73.55 &40.37/61.43 &69.30/71.97	 &68.46/78.74 &53.50/38.25 &61.38/42.84	 &63.76/73.53 &61.84\\
\hline

\textbf{ADDA+MAML}
    &81.33/79.95 &57.63/52.01 &70.10/70.73 &77.16/73.14 &56.81/55.01 &76.40/78.32 &80.69/81.04 &49.34/48.52 &57.84/56.99	 &73.92/72.33 &67.46 \\
\textbf{ADDA+R2D2}
    &79.01/79.15 &57.84/58.43 &71.44/71.82 &78.56/78.43 &58.91/58.74 &77.47/78.13 &82.43/83.02 &49.51/52.41	 &60.70/59.49 &74.12/73.96 &69.18\\
\textbf{ADDA+Meta-ticket}
    &81.46/80.30 &62.03/56.74 &76.45/70.14	 &79.82/78.41 &58.77/52.94 &76.18/75.16	 &81.08/81.51 &59.39/53.22 &55.67/56.25	 &71.43/69.85 &68.84\\
\hline
\textbf{ADDA+ProtoNet} 
    &82.96/81.45 &56.18/54.29 &70.52/71.14	 &79.32/77.35 &52.51/55.76 &75.09/79.47	 &79.38/82.65 &44.36/45.17 &53.81/53.05	 &71.94/70.95 &66.87\\
\textbf{ADDA+DN4} 
    &79.63/71.77 &44.19/59.26 &67.94/65.76	 &64.47/75.64 &39.34/65.72 &72.98/74.77	 &69.26/80.35 &59.21/37.92 &61.29/47.12	 &65.78/79.29 &64.09\\
\textbf{ADDA+ADM} 
    &80.56/72.12 &46.80/61.48 &69.84/65.39	 &68.06/77.28 &49.13/62.59 &74.16/76.42	 &72.12/81.44 &57.95/38.64 &60.85/49.17	 &67.81/79.35 &65.56\\
\textbf{ADDA+FEAT} 
    &81.39/74.36 &46.51/60.13 &68.70/67.65	 &71.94/75.40 &55.19/63.97 &73.27/76.13	 &78.04/80.97 &58.74/45.38 &60.98/54.13	 &69.46/75.62 &66.90\\
\textbf{ADDA+DeepEMD} 
    &80.31/72.65 &47.81/62.25 &67.82/66.74	 &70.45/74.83 &47.61/65.39 &74.59/76.27	 &74.32/81.95 &61.20/46.73 &62.55/50.31	 &67.84/78.54 &66.51\\
\hline
\textbf{IMSE} \cite{huang2021few}
    &73.12/64.71 &39.30/57.28 &63.75/61.39 &64.28/69.30 &33.94/61.03 &68.37/70.07 &66.27/79.01 &52.11/34.12 &59.28/40.02 &60.87/68.47 &59.34\\
\textbf{TSECS} \cite{yu2023high}
    &77.34/79.59 &68.87/65.30 &70.65/70.26 &78.51/79.97 &56.09/64.58 &76.51/89.51 &81.64/83.79 &53.21/56.69 &67.92/75.17 &69.98/73.58 &71.96\\
% \textbf{EMPL-FD}
%     &78.53/70.20 & 57.46/67.82&  66.38/67.93& 73.51/72.69& 46.75/70.18& 69.42/77.74& 73.55/79.63& 55.34/45.76& 68.29/50.98& 65.36/70.46& 66.40\\
% \textbf{EMPL-FD(time)}
%     &(1.53/1.39) & (1.87/1.68)&  (1.72/1.79)& (1.83/1.69)& (1.76/1.84)& (1.89/1.99)& (2.03/1.90)& (1.92/1.76)& (1.74/1.73)& (1.94/1.78)& 1.79\\
\textbf{E$^2$MPL}
    &\textbf{94.80}/\textbf{92.83} &\textbf{78.92}/\textbf{80.12} &\textbf{87.08}/\textbf{90.55} &\textbf{94.46}/\textbf{92.79} &\textbf{82.37}/\textbf{86.72} &\textbf{92.42}/\textbf{94.50} &\textbf{96.97}/\textbf{97.03} &\textbf{73.15}/\textbf{72.14} &\textbf{86.43}/\textbf{80.54} &\textbf{92.55}/\textbf{89.76} &\textbf{87.81}\\
% \textbf{E$^2$MPL(time)}
%     &(\textbf{0.52}/\textbf{0.54}) &\textbf{(0.56/0.57)}  &  \textbf{(0.52/0.55)}& \textbf{(0.58/0.51)}& \textbf{(0.49/0.63)}& \textbf{(0.50/0.57)}& \textbf{(0.57/0.50)}& \textbf{(0.58/0.47)}& \textbf{(0.59/0.52)}& \textbf{(0.59/0.52)}&\textbf{0.54}\\
\hline\hline
\multicolumn{11}{c}{\textbf{5-way, 5-shot, backbone: CLIP (ViT-B/32)}} \\
\hline
\textbf{MCD} \cite{DBLP:conf/cvpr/SaitoWUH18}
     &86.50/73.38 &57.82/64.20 &73.27/74.42 &75.62/75.11 &48.89/62.46 &70.66/76.07 &78.83/86.34 &62.53/38.58 &62.19/46.49 &73.24/80.69 &68.37\\
\textbf{ADDA} \cite{DBLP:conf/cvpr/TzengHSD17}
     &86.62/75.89 &61.81/66.13 &76.25/75.33 &78.99/78.12 &54.63/75.82 &75.38/81.71 &77.57/88.72 &68.37/42.51 &73.65/46.59 &76.92/80.79 &72.09\\
\textbf{DWT}  \cite{roy2019unsupervised}
     &86.91/73.56 &63.26/70.31 &76.85/74.57 &80.78/81.91 &56.80/78.52 &76.57/82.22 &81.06/89.15 &65.96/43.91 &76.74/49.69 &77.08/82.21 &73.41\\
\hline
\textbf{MAML} \cite{DBLP:conf/icml/FinnAL17}
    &92.19/85.51 &56.49/58.15 &80.70/81.07 &88.29/86.46 &52.39/54.10 &84.69/88.42 &89.98/92.44 &45.72/50.76 &59.51/56.84 &84.19/80.15 &73.40\\
\textbf{R2D2} \cite{DBLP:conf/iclr/BertF19}
    &91.52/86.54 &62.30/67.19 &82.24/84.64 &90.55/86.54 &56.56/62.13 &84.70/90.80 &91.06/93.67 &53.40/53.83 &70.48/64.36 &86.86/82.84 &77.11\\
\textbf{Meta-ticket}\cite{DBLP:conf/neurips/ChijYA22}
    &90.18/81.69 &61.07/65.42 &81.70/81.99	 &89.26/86.39 &57.96/56.35 &85.59/89.64	 &89.53/93.26 &51.34/54.89 &60.43/61.82	 &84.24/80.40 &75.16\\
\hline
\textbf{baseline} \cite{chen2019closer}
    &82.17/72.82 &49.41/47.02 &71.33/70.81 &80.69/80.38 &42.96/47.36 &68.50/78.68 &78.33/87.71 &38.70/40.78	 &48.93/48.09 &72.61/72.37 &63.98\\
\textbf{baseline++} \cite{chen2019closer}
    &90.48/85.36 &57.92/66.61 &80.68/80.93 &88.32/85.94 &50.93/65.46 &83.70/86.12 &89.52/92.02 &53.52/45.19	 &67.74/55.97 &81.96/78.70 &74.35\\
\textbf{ProtoNet} \cite{DBLP:journals/corr/SnellSZ17}
    &89.06/83.98 &57.87/61.08 &79.25/82.25 &89.28/86.06 &51.44/55.45 &80.80/89.68 &88.15/90.60 &46.52/50.58	 &63.72/58.36 &83.76/78.61 &73.33\\
\textbf{DN4} \cite{li2019revisiting}
    &89.89/78.62 &63.10/74.23 &79.56/77.88 &81.57/84.79 &40.10/79.77 &80.84/85.79 &80.76/92.85 &67.99/42.82 &77.52/49.15 &78.28/82.13 &74.38\\
\textbf{ADM} \cite{li2020asymmetric}
    &87.40/76.43 &64.57/75.89 &73.05/79.92 &81.98/85.35 &42.02/76.12 &81.90/79.72 &72.26/89.67 &62.91/43.17 &76.84/58.43	 &81.72/83.20 &73.63\\
\textbf{FEAT} \cite{FEAT}
    &89.92/79.64 &67.38/77.60 &81.59/80.23	 &82.37/85.76 &46.10/82.39 &84.97/87.22	 &84.65/93.33 &68.72/48.48 &79.58/57.78	 &83.94/85.81 &77.37\\
\textbf{DeepEMD} \cite{DeepEMD}
    &91.23/82.89 &68.31/77.12 &82.06/81.11 &84.75/84.39 &45.91/83.85 &84.74/89.58 &85.72/92.86 &67.21/49.27 &81.53/56.49	 &83.61/84.95 &77.88\\
\hline
\textbf{ADDA+MAML}
    &93.13/88.58 &66.80/62.32 &82.27/83.69 &89.24/87.00 &57.57/60.76 &85.01/90.39 &90.89/92.29 &46.26/53.32 &67.98/62.25	 &85.65/83.34 &76.44\\
\textbf{ADDA+R2D2}
    &90.60/86.08 &68.29/66.84 &83.20/84.22 &89.77/87.01 &67.08/67.93 &85.72/90.60 &90.46/91.97 &58.61/62.64	 &71.48/68.53 &87.25/83.06 &79.07\\
\textbf{ADDA+Meta-ticket}
    &92.21/82.47 &66.85/69.76 &80.64/82.03	&89.34/86.29 &63.97/67.46 &86.52/89.39	&90.04/93.98 &62.65/61.78 &71.88/71.95	&85.62/80.74 &78.78\\
\hline
\textbf{ADDA+ProtoNet}
    &93.29/85.61 &64.27/66.85 &80.34/83.62	 &91.57/86.73 &52.35/55.64 &82.68/92.87	 &89.36/91.17 &47.29/50.32 &65.42/59.12	 &83.91/79.46 &75.09\\
\textbf{ADDA+DN4}
    &92.75/84.69 &65.23/71.94 &84.31/83.04 &82.70/88.57	 &45.62/79.95 &82.92/90.01	 &84.27/93.49 &68.48/46.91	 &80.15/56.11 &81.93/85.41 &77.42\\
\textbf{ADDA+ADM}
    &92.93/83.14 &66.75/74.45 &81.66/84.72 &83.19/88.87	 &45.95/77.37 &83.57/90.98	 &85.75/93.57 &65.42/49.87	 &78.57/60.21 &82.32/85.53 &77.74\\
\textbf{ADDA+FEAT}
    &94.07/81.99 &71.72/77.53	 &86.74/85.92 &83.68/87.24	 &47.29/84.61 &85.18/91.24	 &87.89/94.01 &69.15/50.39	 &81.33/61.78 &85.19/86.82 &79.69\\
\textbf{ADDA+DeepEMD}
    &93.24/85.96 &70.39/78.29	 &83.17/83.95 &87.36/85.74	 &47.87/84.69 &86.51/91.55	 &86.87/93.17 &67.44/53.52	 &81.97/59.24 &85.42/84.79 &79.56\\
\hline
\textbf{IMSE} \cite{huang2021few}
    &89.70/78.39 &49.58/74.00 &79.55/77.78 &82.37/84.46 &40.45/79.86 &81.06/85.98 &81.04/93.23 &67.71/43.41	 &76.82/49.56 &77.25/81.96 &73.71\\
\textbf{TSECS} \cite{yu2023high}
    &94.78/90.30 &68.38/77.26 &85.55/87.50 &93.78/90.31 &65.91/81.73 &86.43/95.04 &93.46/95.94 &68.33/58.56 &85.66/70.81 &89.25/84.77 &83.19\\
% \textbf{EMPL-FD}
%     &83.90/76.77 & 63.57/71.16&  72.23/73.65& 79.49/76.81& 50.35/74.98& 73.16/81.32& 79.87/83.56& 61.13/49.22& 75.95/56.54& 74.19/73.68& 71.58\\
% \textbf{EMPL-FD(time)}
%     &(1.87/1.96) & (1.96/1.83)&  (1.87/1.82)& (1.89/1.72)& (1.80/1.84)& (1.77/1.87)& (1.85/1.81)& (1.87/2.01)& (1.91/2.01)& (1.87/1.80)& 1.87\\
\textbf{E$^2$MPL}
    &\textbf{98.19}/\textbf{94.76} &\textbf{85.02}/\textbf{89.20} &\textbf{93.19}/\textbf{93.87} &\textbf{97.01}/\textbf{95.33} &\textbf{85.95}/\textbf{93.25} &\textbf{93.15}/\textbf{98.12} &\textbf{97.25}/\textbf{98.01} &\textbf{80.69}/\textbf{82.84} &\textbf{91.95}/\textbf{87.42} &\textbf{96.56}/\textbf{93.10} &\textbf{92.24}\\
% \textbf{EMPL(time)}
%     &\textbf{(0.54/0.42)} &\textbf{(0.56/0.46)}  &  \textbf{(0.52/0.43)}& \textbf{(0.48/0.36)}& \textbf{(0.54/0.48)}& \textbf{(0.39/0.43)}& \textbf{(0.47/0.34)}& \textbf{(0.49/0.52)}& \textbf{(0.47/0.53)}& \textbf{(0.47/0.39)}&\textbf{0.47}\\
% \hline
% \textbf{MELDA-$\theta_P$}
%     &70.73/61.93 &61.73/34.49 &61.89/58.47 &74.35/56.40 &47.78/39.33 &66.04/71.55 &72.11/72.36 &35.16/46.29 &44.58/61.08 &61.04/56.22 &57.68\\
% \textbf{MELDA$\_$CLIP}
%     &89.96/85.90 &67.57/64.08 &80.72/83.00 &89.26/86.63 &67.44/65.11 &82.60/89.10 &90.11/91.16 &54.92/59.41 &66.85/69.69 &85.61/81.36 &77.52 \\
% \textbf{MELDA$\_$CLIP-$\theta_P$}
%     &94.39/88.60 &72.87/65.20 &85.73/86.33 &90.53/88.03 &70.31/72.99 &87.14/91.36 &92.15/94.19 &61.04/64.14 &74.39/73.83 &87.68/83.77 &81.23 \\ 
\hline\hline
\end{tabular}
}

\vspace{-0.3cm}
\end{table*}

\begin{table*}%[t]
\caption{Comparison of our E$^2$MPL and the baseline methods for the FS-UDA setting. The classification accuracy (\%) of target domain samples are averaged over 3600 new tasks under the $5$-way, $K$-shot setting. %The backbone we used is a 12-layer ResNet network (ResNet-12). 
The best performances are shown in bold.}
\label{tab:table22}
\centering
%\scriptsize
\renewcommand{\arraystretch}{1.2}
\setlength{\tabcolsep}{6pt}
% \vspace{-0.4cm}
\resizebox{\linewidth}{!}{
\begin{tabular}{c|cccccccccc|c}
\hline\hline
\multicolumn{11}{c}{\textbf{5-way, 1-shot, backbone: ResNet-12}} \\
\hline \multirow{2}{*}{\textbf{Methods}}
    &\textbf{skt$\leftrightarrow$rel} &\textbf{skt$\leftrightarrow$qdr} &\textbf{skt$\leftrightarrow$pnt} &\textbf{skt$\leftrightarrow$cli} &\textbf{rel$\leftrightarrow$qdr} &\textbf{rel$\leftrightarrow$pnt} &\textbf{rel$\leftrightarrow$cli} &\textbf{qdr$\leftrightarrow$pnt} &\textbf{qdr$\leftrightarrow$cli} &\textbf{pnt$\leftrightarrow$cli} & \textbf{avg}\\
     &$\rightarrow$ / $\leftarrow$&$\rightarrow$ / $\leftarrow$&$\rightarrow$ / $\leftarrow$&$\rightarrow$ / $\leftarrow$&$\rightarrow$ / $\leftarrow$&$\rightarrow$ / $\leftarrow$&$\rightarrow$ / $\leftarrow$&$\rightarrow$ / $\leftarrow$&$\rightarrow$ / $\leftarrow$&$\rightarrow$ / $\leftarrow$ & -- \\\hline
\textbf{MCD} \cite{DBLP:conf/cvpr/SaitoWUH18}
    &48.07/37.74 &38.90/34.51 &39.31/35.59 &51.43/38.98 &24.17/29.85 &43.36/47.32 &44.71/45.68 &26.14/25.02 &42.00/34.69 &39.49/37.28 &38.21\\
\textbf{ADDA} \cite{DBLP:conf/cvpr/TzengHSD17}
    &48.82/46.06 &38.42/40.43 &42.52/39.88 &50.67/47.16 &31.78/35.47 &43.93/45.51 &46.30/47.66 &26.57/27.46 &46.51/32.19 &39.76/41.24 &40.91\\
\textbf{DWT} \cite{roy2019unsupervised}
    &49.43/38.67 &40.94/38.00 &44.73/39.24 &52.02/50.69 &29.82/29.99 &45.81/50.10 &52.43/51.55 &24.33/25.90 &41.47/39.56 &42.55/40.52 &41.38\\
\hline
\textbf{MAML} \cite{DBLP:conf/icml/FinnAL17}
    &43.84/31.19 &25.67/23.12 &32.88/28.69 &33.32/31.69 &23.84/23.08 &39.21/36.14 &35.58/36.35 &22.57/22.95 &25.63/23.51 &29.79/28.97 &29.90\\
\textbf{R2D2} \cite{DBLP:conf/iclr/BertF19}
    &38.48/33.63 &26.16/27.67 &33.47/31.17 &35.92/33.82 &23.80/24.66 &39.65/40.46 &38.49/40.92 &23.39/23.05 &27.36/25.34 &31.01/31.40 &31.49\\
\textbf{Meta-ticket} \cite{DBLP:conf/neurips/ChijYA22}
    &44.12/35.39 &28.79/29.48 &33.39/31.08 &34.85/34.61 &26.28/26.85 &43.64/45.54 &39.85/40.72 &23.31/23.98 &28.27/25.02 &31.48/32.68 &32.97\\
\hline
\textbf{ProtoNet} \cite{DBLP:journals/corr/SnellSZ17}
    &50.48/43.15   &41.20/32.63 &46.33/39.69 &53.45/48.17 &32.48/25.06 &49.06/50.30   &49.98/51.95 &22.55/28.76   &36.93/40.98 &40.13/41.10 &41.21\\
\textbf{DN4} \cite{li2019revisiting}
     &52.42/47.29   &41.46/35.24   &46.64/46.55   &54.10/51.25 &33.41/27.48   &52.90/53.24   &53.84/52.84 &22.82/29.11   &36.88/43.61 &47.42/43.81 &43.61\\
\textbf{ADM} \cite{li2020asymmetric}
      &49.36/42.27 &40.45/30.14 &42.62/36.93 &51.34/46.64 &32.77/24.30 &45.13/51.37 &46.80/50.15 &21.43/30.12 &35.64/43.33 &41.49/40.02
        &40.11 \\
\textbf{FEAT} \cite{FEAT}
     &51.72/45.66   &40.29/35.45   &47.09/42.99   &53.69/50.59  &33.81/27.58   &52.74/53.82   &53.21/53.31 &23.10/29.39   &37.27/42.54 &44.15/44.49 &43.14\\
\textbf{DeepEMD} \cite{DeepEMD}
     &52.24/46.84   &42.12/34.77   &46.64/43.89   &55.10/49.56 &34.28/28.02   &52.73/53.26   &54.25/54.91 &22.86/28.79   &37.65/42.92 &44.11/44.38 &43.46\\
\hline
\textbf{ADDA+MAML}
    &41.01/31.23 &24.30/23.56 &30.14/28.16 &33.46/30.43	&22.93/25.64 &38.87/36.10 &37.19/38.72 &23.98/25.46 &26.72/28.97 &31.16/31.96 &30.50\\
\textbf{ADDA+R2D2}
    &36.51/33.57 &23.73/23.35 &31.16/29.55 &33.55/32.83	&23.67/22.42 &39.80/37.69 &39.31/38.90 &22.11/21.90	&23.57/23.14 &30.23/31.98 &29.95\\
\textbf{ADDA+Meta-ticket}
    &43.28/36.38 &25.71/24.64 &33.41/31.27 &38.81/33.97 &25.62/26.74 &45.33/44.21 &40.54/41.68 &24.79/25.16 &28.29/25.62 &31.93/32.85 &33.01\\
\hline
\textbf{ADDA+ProtoNet}
    &51.30/43.43 &41.79/35.40 &46.02/41.40 &52.68/48.91 &37.28/27.68 &50.04/49.68 &49.83/52.58 &23.72/32.03 &38.54/44.14 &41.06/41.59 &42.45 \\
\textbf{ADDA+DN4}
    &53.04/46.08 &42.64/36.46 &46.38/47.08 &54.97/51.28 &34.80/29.84 &	53.09/54.05 &54.81/55.08 &23.67/31.62 &42.24/45.24 &46.25/44.40 &44.65\\
\textbf{ADDA+ADM}
      &51.87/45.08  &43.91/32.38  &47.48/43.37  &54.81/51.14  &35.86/28.15  &48.88/51.61  &49.95/54.29  &23.95/33.30  &43.59/48.21  &43.52/43.83 & 43.76\\
\textbf{ADDA+FEAT}
    &52.72/46.08 &47.00/36.94 &47.77/45.01 &56.77/52.10 &36.32/30.50 &49.14/52.36 &52.91/53.86 &24.76/35.38 &44.66/48.82 &45.03/45.92 &45.20\\
\textbf{ADDA+DeepEMD}
    &53.98/47.55 &44.64/36.19 &46.29/45.14 &55.93/50.45  &37.47/30.14 &52.21/53.32 &54.86/54.80  &23.46/32.89 &39.06/46.76  &45.39/44.65 &44.75\\
\hline
\textbf{IMSE} \cite{huang2021few}
    &57.21/51.30 &49.71/40.91 &50.36/46.35 &59.44/54.06 &44.43/36.55 &52.98/55.06 &57.09/57.98 &30.73/38.70 &48.94/51.47 &47.42/46.52 &48.86\\
\textbf{TSECS} \cite{yu2023high}
    &65.00/58.22 &62.25/51.97 &56.51/53.70 &69.45/64.59 &56.66/\textbf{49.82} &58.76/63.18 &67.98/67.89 &\textbf{38.26}/46.15 &\textbf{60.51}/\textbf{69.03} &54.40/52.76 &58.20\\
\textbf{E$^2$MPL}
    &\textbf{67.13}/\textbf{66.56} &\textbf{62.51}/\textbf{52.82} &\textbf{59.61}/\textbf{61.09} &\textbf{72.35}/\textbf{67.17} &\textbf{57.41}/46.32 &\textbf{65.92}/\textbf{68.60} &\textbf{69.39}/\textbf{73.83} &36.19/\textbf{49.15} &59.04/66.29 &\textbf{62.67}/\textbf{61.17} &\textbf{61.26}\\
% \hline
% \textbf{MELDA-$\theta_P$}
%     &55.55/52.10 &49.13/33.51 &47.98/46.52 &57.81/52.83 &43.47/37.09 &51.92/55.85 &58.94/56.35 &28.13/36.27 &45.53/53.10 &47.61/44.80 &47.72\\
% \textbf{MELDA$\_$CLIP}
%     &78.98/79.09 &60.05/58.16 &71.18/71.87 &79.00/78.26 &60.70/57.30 &76.06/76.53 &81.93/82.34 &48.83/52.15 &59.53/62.25 &74.63/72.68 &69.08\\
% \textbf{MELDA$\_$CLIP-$\theta_P$}
%     &89.74/87.67 &68.73/62.84 &80.45/82.11 &88.69/87.89 &69.64/65.69 &85.57/89.47 &92.02/92.91 &55.15/62.32 &65.94/73.76 &86.71/82.88 &78.51\\ 

\hline\hline
\multicolumn{11}{c}{\textbf{5-way, 5-shot, backbone: ResNet-12}} \\
\hline 
\textbf{MCD} \cite{DBLP:conf/cvpr/SaitoWUH18}
    &66.42/47.73 &51.84/39.73 &54.63/47.75 &72.17/53.23 &28.02/33.98 &55.74/66.43 &56.80/63.07 &28.71/29.17 &50.46/45.02 &53.99/48.24 &49.65\\
\textbf{ADDA} \cite{DBLP:conf/cvpr/TzengHSD17}
    &66.46/56.66 &51.37/42.33 &56.61/53.95 &69.57/65.81 &35.94/36.87 &58.11/63.56 &59.16/65.77 &23.16/33.50 &41.94/43.40 &55.21/55.86 &51.76\\
\textbf{DWT} \cite{roy2019unsupervised}
    &67.75/54.85 &48.59/40.98 &55.40/50.64 &69.87/59.33 &36.19/36.45 &60.26/68.72 &62.92/67.28 &22.64/32.34 &47.88/50.47 &49.76/52.52 &51.74\\
\hline
\textbf{MAML} \cite{DBLP:conf/icml/FinnAL17}
    &52.59/38.76 &28.04/28.42 &38.46/34.98 &43.28/40.45 &27.52/27.36 &50.49/52.71 &44.63/53.86 &23.81/24.25 &28.92/29.31 &38.11/37.73 &37.18\\
\textbf{R2D2} \cite{DBLP:conf/iclr/BertF19}
    &50.51/39.78 &29.31/28.78 &39.48/36.88 &44.07/40.44 &29.43/27.81 &50.51/53.37 &46.62/54.62 &24.79/26.53 &31.63/32.15 &39.07/39.54 &38.27\\
\textbf{Meta-ticket} \cite{DBLP:conf/neurips/ChijYA22}
    &54.62/41.38 &30.72/31.47 &43.82/38.69 &46.12/41.52 &30.32/29.17 &51.60/54.54 &47.28/56.36 &25.78/26.93 &32.78/34.17 &40.12/41.77 &39.99\\
\hline
\textbf{ProtoNet} \cite{DBLP:journals/corr/SnellSZ17}
     &65.07/56.21  &52.65/39.75  &55.13/52.77   &65.43/62.62
     &37.77/31.01   &61.73/66.85   &63.52/66.45
     &20.74/30.55  &45.49/55.86
     &53.60/52.92 &51.80\\
\textbf{DN4} \cite{li2019revisiting}
     &63.89/51.96   &48.23/38.68   &52.57/51.62   &62.88/58.33 &37.25/29.56   &58.03/64.72   &61.10/62.25 &23.86/33.03   &41.77/49.46 &50.63/48.56 &49.41\\
\textbf{ADM} \cite{li2020asymmetric}
    &66.25/54.20 &53.15/35.69 &57.39/55.60 &71.73/63.42 &44.61/24.83 &59.48/69.17 &62.54/67.39 &21.13/38.83 &42.74/58.36 &56.34/52.83 &52.78\\
\textbf{FEAT} \cite{FEAT}
     &67.91/58.56 &52.27/40.97 &59.01/55.44 &69.37/65.95 &40.71/28.65 &63.85/71.25 &65.76/68.96 &23.73/34.02 &42.84/53.56 &57.95/54.84 &53.78\\
\textbf{DeepEMD} \cite{DeepEMD}
     &67.96/58.11 &53.34/39.70 &59.31/56.60 &70.56/64.60 &39.70/29.95 &62.99/70.93 &65.07/69.06 &23.86/34.34 &45.48/53.93 &57.60/55.61 &53.93\\
\hline
\textbf{ADDA+MAML}
    &52.56/39.25 &29.99/27.63 &40.89/37.91 &45.40/45.21 &27.58/27.84 &50.63/53.30 &48.77/50.46 &23.11/22.18 &27.12/26.69 &37.55/40.83 &37.75\\
\textbf{ADDA+R2D2}
    &52.15/41.34 &30.13/26.48 &44.76/38.33 &46.97/45.44 &26.79/27.74 &49.87/50.82 &49.98/51.69 &23.76/23.52 &30.75/30.07 &39.72/42.03 &38.12\\
\textbf{ADDA+Meta-ticket}
    &56.31/42.70 &31.52/32.22 &43.64/40.14 &48.97/47.95 &28.64/29.37 &54.39/57.69 &53.73/56.91 &25.85/24.62 &32.24/30.10 &40.67/41.17 &40.94\\
\hline
\textbf{ADDA+ProtoNet}
    &66.11/58.72 &52.92/43.60 &57.23/53.90 &68.44/61.84 &45.59/38.77 &60.94/69.47 &66.30/66.10 &25.45/41.30 &46.67/56.22 &58.20/52.65 &54.52\\
\textbf{ADDA+DN4}
    &63.40/52.40 &48.37/40.12 &53.51/49.69 &64.93/58.39 &36.92/31.03 &57.08/65.92 &60.74/63.13 &25.36/34.23 &48.52/51.19 &52.16/49.62 &50.33\\
\textbf{ADDA+ADM}
      &64.64/54.65  &52.56/33.42  &56.33/54.85  &70.70/63.57  &39.93/27.17  &58.63/68.70  &61.96/67.29  &21.91/39.12  &41.96/59.03  &55.57/53.39 & 52.27\\
\textbf{ADDA+FEAT}
    &67.80/56.71 &60.33/43.34 &57.32/58.08 &70.06/64.57 &44.13/35.62 &62.09/70.32 &57.46/67.77 &29.08/44.15 &49.62/63.38 &57.34/52.13 &55.56\\

\textbf{ADDA+DeepEMD}
    &68.52/59.28 &56.78/40.03 &58.18/57.86 &70.83/65.39
    &42.63/32.18 &63.82/71.54 &66.51/69.21
    &26.89/42.33 &47.00/57.92
    &57.81/55.23 &55.49\\
\hline
\textbf{IMSE} \cite{huang2021few}
    &70.46/61.09 &61.57/46.86 &62.30/59.15 &76.13/67.27 &53.07/40.17 &64.41/70.63 &67.60/71.76 &33.44/48.89 &53.38/65.90 &61.28/56.74 &59.60\\
\textbf{TSECS} \cite{yu2023high}
    &78.23/\textbf{70.44} &\textbf{77.90}/55.77 &66.70/68.03 &83.82/74.28 &64.33/\textbf{55.16} &68.40/79.74 &78.23/77.69 &39.74/63.02 &\textbf{67.99}/\textbf{80.31} &73.67/61.63 &69.25\\
\textbf{E$^2$MPL}
    &\textbf{80.28}/69.77 &77.61/\textbf{60.64} &\textbf{71.27}/\textbf{72.34} &\textbf{86.00}/\textbf{76.34} &\textbf{69.10}/53.86 &\textbf{72.37}/\textbf{84.71} &\textbf{83.15}/\textbf{85.03} &\textbf{44.89}/\textbf{64.32} &62.77/78.54 &\textbf{77.52}/\textbf{71.38} &\textbf{71.90}\\
% \hline
% \textbf{MELDA-$\theta_P$}
%     &70.73/61.93 &61.73/34.49 &61.89/58.47 &74.35/56.40 &47.78/39.33 &66.04/71.55 &72.11/72.36 &35.16/46.29 &44.58/61.08 &61.04/56.22 &57.68\\
% \textbf{MELDA$\_$CLIP}
%     &89.96/85.90 &67.57/64.08 &80.72/83.00 &89.26/86.63 &67.44/65.11 &82.60/89.10 &90.11/91.16 &54.92/59.41 &66.85/69.69 &85.61/81.36 &77.52 \\
% \textbf{MELDA$\_$CLIP-$\theta_P$}
%     &94.39/88.60 &72.87/65.20 &85.73/86.33 &90.53/88.03 &70.31/72.99 &87.14/91.36 &92.15/94.19 &61.04/64.14 &74.39/73.83 &87.68/83.77 &81.23 \\ 
\hline\hline
\end{tabular}
}

\vspace{-0.3cm}
\end{table*}

\subsection{Comparison Methods and Results}
We compare our E$^2$MPL with twenty-three related methods on \emph{DomainNet} in Table \ref{tab:table222}. This includes three popular UDA methods (\emph{i.e.}, MCD \cite{DBLP:conf/cvpr/SaitoWUH18}, ADDA \cite{DBLP:conf/cvpr/TzengHSD17} and DWT \cite{roy2019unsupervised}), three meta-learning methods (\emph{i.e.}, MAML \cite{DBLP:conf/icml/FinnAL17}, R2D2 \cite{DBLP:conf/iclr/BertF19} and Meta-ticket \cite{DBLP:conf/neurips/ChijYA22}), seven FSL methods (\emph{i.e.}, baseline/baseline++ \cite{chen2019closer}, ProtoNet \cite{DBLP:journals/corr/SnellSZ17}, DN4 \cite{li2019revisiting}, ADM \cite{li2020asymmetric}, FEAT \cite{FEAT} and DeepEMD \cite{DeepEMD}), as well as two FS-UDA methods (\emph{i.e.}, IMSE \cite{huang2021few} and TSECS \cite{yu2023high}). Then, we evaluate variants of meta-learning methods by integrating ADDA for domain adaptation \cite{DBLP:conf/cvpr/TzengHSD17}, \emph{i.e.,} ADDA+MAML, ADDA+R2D2 and ADDA+Meta-ticket. In addition, we also combine the aforementioned five FSL methods with ADDA \cite{DBLP:conf/cvpr/TzengHSD17}, which are abbreviated as ADDA+ProtoNet, ADDA+DN4, ADDA+ADM, ADDA+FEAT, and ADDA+DeepEMD, respectively. For most of the methods involved in this comparison, we utilize their publicly available open-source codes and configure their optimal parameters for implementation. Furthermore, for fair comparison, all comparison experiments are conducted on the same data, and we modify these compared methods for the FS-UDA setting as follows.

% {\bf For the UDA methods of DANN, ADDA and MCD,} we first train a UDA model on the whole auxiliary dataset, and then finetune the fully-connected layers of the learned UDA model to adapt to new FS-UDA tasks to test. Thus, they are called `DANN+finetune', `ADDA+finetune' and `MCD+finetune’, respectively. 
% To improve the task-level generalization of existing UDA models, 
% %to compare with the proposed MELDA, 
% we specially design {\bf the method `MCD+R2-D2'}. It first trains an MCD model and then leverages the R2-D2 method in \cite{DBLP:conf/iclr/BertF19} to train a meta learner, based on feature embeddings learned by MCD, for generalization. 
{\bf For the UDA methods of MCD, ADDA and DWT,} we first train a UDA model on the whole auxiliary dataset. Once the model is trained, we proceed to adapt the model to new FS-UDA tasks by fine-tuning the fully connected layers, ensuring that it performs well on the new task during testing. 
%Thus, they are called `MCD+finetune', `ADDA+finetune' and `DWT+finetune’, respectively. 
%\re{Since we did not design these methods for the few-shot samples, these methods may be affected by the lack of samples from the source domain.}
{\bf For the meta-learning methods of MAML, R2D2 and Meta-ticket,} we directly train a model over meta-tasks by only leveraging their source domain data for few-shot classification to optimize the model parameters, and then migrate the models to the target domain to evaluate the adaptability of model on the new task in the meta-testing phase. %\re{They did not specially handle domain shift and this could affect their performance. In addition, multi-step updates in meta-learning and some similarity calculation will lead to time-consuming network training.}
% To modify {\bf the few-shot SDA method FADA} for our setting, we first leverage auxiliary source domain data to pretrain the backbone network, then train a FADA model on the whole auxiliary data, and finally finetune the model to adapt to new FS-UDA tasks. Note that, since FADA needs target domain labels to calculate the domain adversarial loss, we specially leverage five labeled samples per category in target domain of the auxiliary data to train the FADA model, which is different from our E$^2$MPL without using any target labels.
{\bf For the few-shot learning methods of baseline, baseline++, ProtoNet, DN4, ADM, FEAT and DeepEMD} applied for our setting, we train a pre-trained model on the source domain data, and then the pre-trained model is adapted to the target domain data. %we train a model using similarity measures for each class prototype constructed by the source domain support set and labeled source domain query images. Then, the trained model is applied to the query set of the target domain to test the performance of model on the FS-UDA task.
{\bf For the eight combination methods}, taking ADDA+MAML as an example, we pre-train the MAML on the source domain to learn effective feature representations. Simultaneously, we employ an unsupervised domain adaptation strategy like Adversarial Discriminative Domain Adaptation (ADDA) to align the source and target domains. %\re{We compare it with our E$^2$MPL, which learns a meta learner only for feature embedding and uses ridge regression to build task-specific adapter and classifiers.} % As seen in Table \ref{tab:table22} and \ref{tab:table222}, calling it MAML-DA, we tested it on different backbones (ResNet-12 and CLIP), especially on CLIP, and it already performed very well.
Next, {\bf for the FS-UDA methods of IMSE and TSECS,} we can directly evaluate them since their experimental setting is the same as ours. %\re{The difference between the three methods is that IMSE and TSECS are metric-based methods, while E$^2$MPL is optimization-based methods. In addition, }
%\re{The two methods use local features, which have high requirements for learning key discriminative features, while our E$^2$MPL is designed to learn a good initialization parameter and quickly learn image semantic knowledge of new tasks.}

%Finally, we propose \re{an extensive work {\bf E$^2$MPL}}. \re{In this method, we introduce a pre-trained model and freeze its parameters, adding some learnable virtual tokens as prompts to it for fine-tuning. We focus on optimizing the prompt module. The prompts are used to store the meta-knowledge learned by the above mentioned three losses. This methods not only helps to preserve the general knowledge stored by the large model, but also significantly reduces the number of training parameters and adapts the model efficiently to various downstream tasks. }

%\textbf{Comparison Results.}
%We conduct comparison on \emph{DomainNet} for $5$-way $1$-shot and $5$-shot UDA tasks. 
%For backbones, the baselines apply four-layer convolutional networks or ResNet-12 to build the feature embedding model.
To better demonstrate the advantages of the E$^2$MPL, we first use the pre-trained model CLIP as the backbone, and then report the average classification accuracy of target domain samples over 3600 tasks tested. 
%As seen in Table \ref{tab:table2}, our MELDA achieves the best performance for either backbone, 
% As seen in Table \ref{tab:table2}, our MELDA \yl{and MELDA-$\theta_p$} consistently achieves better performance than the compared ones on the two datasets for all three backbones. This is true for both 1-shot and 5-shot tasks in either of the cross-domain ways.
As seen in Table \ref{tab:table222}, our E$^2$MPL consistently and significantly achieves the best performance with only a few parameters learned, compared to the other methods on \emph{DomainNet} for both 1-shot and 5-shot tasks. It shows the effectiveness of our E$^2$MPL method for FS-UDA. To verify that our method is not only applicable to large pre-trained models (CLIP), but also works well on other traditional models, \emph{i.e.,} ResNet-12. As shown in Table \ref{tab:table22}, using ResNet-12 as the backbone, we compare E$^2$MPL with twenty-one related methods on \emph{DomainNet}: three UDA methods (\emph{i.e.}, MCD, ADDA and DWT), three meta-learning methods (\emph{i.e.}, MAML, R2D2, Meta-ticket), and five popular few-shot methods (\emph{i.e.}, ProtoNet, DN4, ADM, FEAT and DeepEMD). We also combine the few-shot methods with ADDA, \emph{e.g.,} ADDA+MAML, ADDA+R2D2. Meanwhile, we also compare and analyze E$^2$MPL with IMSE and TSECS designed specifically for FS-UDA.

% Moreover, \re{to train with fewer parameters for easier optimization and to better exert the advantages of prompt learning in E$^2$MPL},
%To verify that our method is not only applicable to large pre-trained models(CLIP), but also works well on other traditional models, we also used ResNet-12 as our backbone for comparison. As shown in Table \ref{tab:table22},
% we compare E$^2$MPL with several related methods on \emph{DomainNet}: three UDA methods (\emph{i.e.}, MCD, ADDA and DWT), five popular few-shot methods (\emph{i.e.}, ProtoNet, DN4, ADM, FEAT and DeepEMD), and three meta-learning methods (\emph{i.e.}, MAML, R2D2, Meta-ticket). We also combine the few-shot methods with ADDA, which are abbreviated as \re{ADDA+MAML, ADDA+R2D2 et al.}, respectively. Meanwhile, in the case of ResNet-12 as backbone, we will also compare and analyze E$^2$MPL with IMSE and TSECS designed specifically for FS-UDA. 

% All methods based on CLIP have greatly improved performance \re{over those based on ResNet-12}, especially about meta-learning methods, which also shows powerful representation ability and better generalization ability.%on the basis of \re{the original}
% Whether based on CLIP or ResNet-12, E$^2$MPL has a great improvement over other FS-UDA methods, especially in terms of meta-learning methods, showing strong representation and better generalization ability.
\textbf{In general, whether based on CLIP or ResNet-12, E$^2$MPL has a great improvement over other methods. This is due to the fact that our meta-prompt learning framework can learn meta-prompts shared between tasks as well as task-specific prompts to quickly adapt to new tasks, thus improving the model generalizability.}

\begin{table*}
\centering
\caption{Comparison of related prompt learning methods. `*' indicates that its prompt module replaces our prompts in our E$^2$MLP meta-prompt learning framework as it cannot be used directly. }
\label{tab:tablet}
% \vspace{-0.1cm}
\renewcommand{\arraystretch}{1.2}
\setlength{\tabcolsep}{6pt}
\resizebox{\linewidth}{!}{
\begin{tabular}{c|cccccccccc|c}
\hline\hline
\multicolumn{11}{c}{\textbf{5-way, 1-shot, backbone: CLIP(ViT-B/32)}} \\
\hline \multirow{2}{*}{\textbf{Methods}}
&\textbf{skt$\leftrightarrow$rel} &\textbf{skt$\leftrightarrow$qdr} &\textbf{skt$\leftrightarrow$pnt} &\textbf{skt$\leftrightarrow$cli} &\textbf{rel$\leftrightarrow$qdr} &\textbf{rel$\leftrightarrow$pnt} &\textbf{rel$\leftrightarrow$cli} &\textbf{qdr$\leftrightarrow$pnt} &\textbf{qdr$\leftrightarrow$cli} &\textbf{pnt$\leftrightarrow$cli} & \textbf{avg}\\
     &$\rightarrow$ / $\leftarrow$&$\rightarrow$ / $\leftarrow$&$\rightarrow$ / $\leftarrow$&$\rightarrow$ / $\leftarrow$&$\rightarrow$ / $\leftarrow$&$\rightarrow$ / $\leftarrow$&$\rightarrow$ / $\leftarrow$&$\rightarrow$ / $\leftarrow$&$\rightarrow$ / $\leftarrow$&$\rightarrow$ / $\leftarrow$ & -- \\\hline
\textbf{DAM-VP}
     % &91.75/88.67 &71.57/75.40 &82.44/84.24 &90.62/88.30	 &75.60/79.62 &86.45/90.36 &93.18/94.27 &65.66/66.52	 &81.38/76.88 &87.64/84.65 &82.76\\
     &81.79/78.99 &56.74/57.25 &73.19/74.64	 &81.52/81.59 &52.73/55.68 &77.64/81.28	 &83.69/85.92 &47.30/47.18 &59.24/55.96	 &76.44/75.66 &69.22\\
% \textbf{DAM-VP}*
     % &91.75/88.67 &71.57/75.40 &82.44/84.24 &90.62/88.30	 &75.60/79.62 &86.45/90.36 &93.18/94.27 &65.66/66.52	 &81.38/76.88 &87.64/84.65 &82.76\\
\textbf{ProMetaR}*
     &93.45/89.81 &68.52/76.40 &83.81/86.01 &87.89/90.06	 &69.75/81.77 &89.00/92.78 &93.63/94.66 &66.79/65.42	 &81.53/74.49 &88.67/86.24 &83.03\\
\textbf{PGN}*
     &92.52/90.38 &73.47/77.35 &84.95/86.57 &91.44/90.08 &73.44/81.59 &88.59/92.22 &94.06/95.03 &69.03/66.01 &83.58/77.46 &89.10/87.30 &84.21\\
\textbf{VPT}*
     &93.77/91.24 &74.53/78.56 &86.90/87.84 &93.13/91.03	 &77.78/82.43 &89.50/92.89 &94.11/95.55 &70.43/69.52	 &83.82/78.75 &89.86/88.35 &85.50\\

\textbf{E$^2$MPL}
    &\textbf{94.80}/\textbf{92.83} &\textbf{78.92}/\textbf{80.12} &\textbf{87.08}/\textbf{90.55} &\textbf{94.46}/\textbf{92.79} &\textbf{82.37}/\textbf{86.72} &\textbf{92.42}/\textbf{94.50} &\textbf{96.97}/\textbf{97.03} &\textbf{73.15}/\textbf{72.14} &\textbf{86.43}/\textbf{80.54} &\textbf{92.55}/\textbf{89.76} &\textbf{87.81}\\
\hline\hline
\multicolumn{11}{c}{\textbf{5-way, 5-shot, backbone: CLIP (ViT-B/32)}} \\
\hline
\textbf{DAM-VP}
    &90.33/84.22 &60.12/62.56 &81.78/83.46	 &89.66/83.72 &59.87/64.61 &81.55/90.81	 &87.12/91.05 &65.76/62.64 &68.92/58.70	 &86.61/79.73 &76.66 \\
% \textbf{DAM-VP}*
%      &94.93/91.65 &76.21/80.77 &88.30/89.58 &93.40/92.78	 &76.01/85.63 &90.66/94.33 &94.68/96.57 &70.18/74.02	 &86.39/78.51 &91.45/90.56 &86.83\\
\textbf{ProMetaR}*
     &95.39/91.50 &79.47/79.75 &88.30/89.67 &94.10/91.74	 &77.32/85.73 &90.22/95.27 &95.03/96.28 &72.70/72.33	 &87.75/80.13 &92.45/88.11 &87.16\\
\textbf{PGN}*
     &96.37/92.26 &81.63/81.78 &90.00/91.26 &94.81/92.57 &80.07/85.51 &91.01/96.09 &95.50/96.92 &73.85/76.09 &85.66/81.72 &94.35/90.55 &88.40\\
\textbf{VPT}*
     &96.74/93.13 &82.74/82.64 &90.58/92.24 &95.27/93.57	 &81.28/89.29 &91.97/96.59 &96.15/97.07 &74.95/77.81	 &88.33/83.22 &94.82/90.94 &89.46\\
\textbf{E$^2$MPL}
    &\textbf{98.19}/\textbf{94.76} &\textbf{85.02}/\textbf{89.20} &\textbf{93.19}/\textbf{93.87} &\textbf{97.01}/\textbf{95.33} &\textbf{85.95}/\textbf{93.25} &\textbf{93.15}/\textbf{98.12} &\textbf{97.25}/\textbf{98.01} &\textbf{80.69}/\textbf{82.84} &\textbf{91.95}/\textbf{87.42} &\textbf{96.56}/\textbf{93.10} &\textbf{92.24}\\
% \multirow{2}{*}{\textbf{Methods}} &
% \multicolumn{2}{c}{\textbf{all(-skt) $\Rightarrow$ skt}} && \multicolumn{2}{c}{\textbf{all(-rel) $\Rightarrow$ rel}}&&\multicolumn{2}{c}{\textbf{all(-qdr) $\Rightarrow$ qdr}} && \multicolumn{2}{c}{\textbf{all(-pnt) $\Rightarrow$ pnt}} && \multicolumn{2}{c}{\textbf{all(-cli) $\Rightarrow$ cli}}\\\cline{2-3}\cline{5-6}\cline{8-9}\cline{11-12}\cline{14-15}
% & 1-shot&5-shot&&1-shot&5-shot&&1-shot&5-shot&&1-shot&5-shot&&1-shot&5-shot\\
% \hline
% \textbf{VPT \cite{vpt}}& 86.21&89.47   && 90.26&93.72  && 71.81&79.88   && 82.02&86.35  && 89.76 &92.58  \\
% \textbf{PGN \cite{PGN}}& 84.16&88.05   && 88.63&92.65  && 68.72&76.51  && 79.45&84.48 && 88.21&92.16   \\ 
% \textbf{DAM-VP \cite{DAM-VP}}&   85.62&88.68   && 90.21&92.74  && 72.27&80.15   && 82.11&87.43 && 88.81&92.41   \\
% \textbf{ProMetaR \cite{ProMetaR}} & 84.87&88.34     && 90.05&93.17   && 71.61 &81.36 && 80.22&88.93 && 87.90 &92.99\\
% \textbf{E$^2$MPL}&   \textbf{89.07}&\textbf{93.29}   && \textbf{93.26} &\textbf{96.89}  && \textbf{78.49} &\textbf{85.31}   && \textbf{85.60} &\textbf{90.03} && \textbf{92.60} &\textbf{95.69} \\

\hline\hline
\end{tabular}
}

%\vspace{-0.3cm}
\end{table*}

\begin{table*}
\centering
\caption{Ablation study of the domain-shared and task-specific prompts designed in our E$^2$MPL (backbone CLIP), where \textbf{all (-skt) $\Rightarrow$ skt} indicates the average accuracy when other domains except for \emph{sketch} are the source domain and \emph{sketch} is the target domain.}
%\vspace{-0.1cm}
\label{tab:tablep}
\renewcommand{\arraystretch}{1.2}
\setlength{\tabcolsep}{5pt}
\begin{tabular}{cc|cccccccccccccc}
\hline
\multirow{2}{*}{\textbf{domain-shared}} & \multirow{2}{*}{\textbf{task-specific}} &%\multirow{2}{*}{$\gamma_d^{*}/\gamma_\omega^{*}$} & 
% % \multicolumn{2}{c|}{\emph{miniImageNet*}} &
% % \multicolumn{2}{c}{\emph{Office-Home}} \\\cline{3-6}%\cline{7-8}
% % &&\textbf{Ph $\Rightarrow$ Sk} &
% % \textbf{Sk $\Rightarrow$ Ph} &
% % \textbf{Cl $\Rightarrow$ Pr} &
% % \textbf{Pr $\Rightarrow$ Cl} \\
\multicolumn{2}{c}{\textbf{all(-skt) $\Rightarrow$ skt}} && \multicolumn{2}{c}{\textbf{all(-rel) $\Rightarrow$ rel}}&&\multicolumn{2}{c}{\textbf{all(-qdr) $\Rightarrow$ qdr}} && \multicolumn{2}{c}{\textbf{all(-pnt) $\Rightarrow$ pnt}} && \multicolumn{2}{c}{\textbf{all(-cli) $\Rightarrow$ cli}}\\\cline{3-4}\cline{6-7}\cline{9-10}\cline{12-13}\cline{15-16}
&& 1-shot&5-shot&&1-shot&5-shot&&1-shot&5-shot&&1-shot&5-shot&&1-shot&5-shot\\
\hline
& & 83.93&88.10   && 87.92&92.77   & & 71.13&77.43   && 79.17&84.91  && 87.22&92.25   \\
\checkmark& & 88.05&91.89   && 92.04&95.81  && 76.57&83.73   && 84.38&89.08 && 91.03&94.67   \\ %86.21&89.47   && 90.26&93.72  && 71.81&79.88   && 82.02&86.35  && 89.76 &92.58  \\
&\checkmark &   87.26&87.34   && \textbf{93.79}&94.97  && 76.72&81.89   && 82.65&87.86 && 89.64&92.91   \\
\checkmark &\checkmark &   \textbf{89.07}&\textbf{93.29}   && 93.26 &\textbf{96.89}  && \textbf{78.49} &\textbf{85.31}   && \textbf{85.60} &\textbf{90.03} && \textbf{92.60} &\textbf{95.69} \\
% %\checkmark & \checkmark &  &  33.77/41.05   & 34.85/42.73     & 32.86/42.08  & \textbf{34.15}/40.14  \\
% % \checkmark & \checkmark&& 88.05&92.83 &&92.03& 96.53 &&76.57&85.17&&84.38&89.80&& 91.03&95.02\\
% % \checkmark & \checkmark& \checkmark& \textbf{89.07}&\textbf{93.29}&&\textbf{93.26}&\textbf{96.89}&&\textbf{78.49}&\textbf{85.31}&&\textbf{85.60}&\textbf{90.03}&& \textbf{92.60}&\textbf{95.69}\\
\hline
\end{tabular}
% %\vspace{-0.3cm}
\end{table*}

\textbf{Comparison analysis between our E$^2$MPL and other methods.} Through the observation and analysis, it is found that the reason why the UDA, meta learning and FSL methods are not effective is that the UDA models are likely to overfit the limited labeled samples, and the meta learning and FSL methods fail to deal with the issue of domain gaps. Even traditional meta-learning method combined with UDA methods (ADDA+MAML) may not be able to effectively deal with the FS-UDA problem, because the features and distributions of the source data can interfere with learning in the target domain, making it difficult for the model to extract effective, representative features. As a result, the model may become confined to the limitations of the training data, leading to overfitting. Therefore, simply combining these two approaches is insufficient to tackle the complex challenges, necessitating more refined strategies to address the dual issues of task generalization and domain gaps. E$^2$MPL fixed the pre-trained model, preserving the model's ability to learn general features. Additionally, through meta-prompt learning strategies, the model gained the capability to learn domain-shared features as well as task-specific discriminative features, significantly reducing the risk of model overfitting. 

% The goal of FSL (baseline, baseline++, ProtoNet \emph{et al.}) is to improve the generalization performance of the model by making full use of the limited labeled sample. However, the FSL method alone can not directly address the issues as it overlooks the gaps between different domains. Traditional meta-learning method combined with UDA methods (ADDA+MAML) may not be able to effectively deal with the problem that the generalization of model performance is still limited in the target domain, because the domain gaps may be large, and the methods may not be able to overcome this bias. 

\textbf{Comparison analysis between the two backbones CLIP and ResNet.}
Also, it is obvious that using CLIP as a backbone can better leverage the advantages of E$^2$MPL, resulting in more significant improvements compared to ResNet. This may be due to CLIP pre-trained on large-scale datasets, which provides it with richer semantic knowledge. By optimizing the prompts alone, the model retains the ability to learn general features while also acquiring the key characteristics of new tasks. Nevertheless, using ResNet as a backbone, E$^2$MPL still performs better on the FS-UDA problem than methods like IMSE and TSECS. This is because the meta-prompt framework provides prompt enhancements for new tasks, where prompts learned from previous tasks can assist in the learning of new tasks, thereby guiding the approach.

\subsection{Evaluation of Prompt Learning}

\begin{table*}
\centering
%\small
%\scriptsize
%\vspace{-0.2cm}
\caption{The effect of domain generalization in our E$^2$MPL: the averaged classification accuracy (\%) and variance over 3600 tasks on \emph{DomainNet} for backbone CLIP. The higher accuracy and less variance value are shown in bold.} %varying combinations are evaluated on the testing sets of \emph{miniImageNet} (left) and \emph{Office-Home} (bottom). `64-64-64-64'MELDA-$\theta_p$
%The averaged accuracy values of target domain over 100 tasks under the 5-way, 1-shot or 5-shot setting are reported in percent, and the best performances are in bold. }
%\vspace{-0.1cm}
\label{tab:table5}
\renewcommand{\arraystretch}{1.2}
\setlength{\tabcolsep}{5pt}
\begin{tabular}{c|ccccccccc}
\hline
\hline
meta-train & skt $\Rightarrow$ rel & skt $\Rightarrow$ rel & skt $\Rightarrow$ rel & pnt $\Rightarrow$ cli & pnt $\Rightarrow$ cli & pnt $\Rightarrow$ cli & qdr $\Rightarrow$ pnt& qdr $\Rightarrow$ pnt& qdr $\Rightarrow$ pnt\\
 meta-test & skt $\Rightarrow$ qdr & skt $\Rightarrow$ pnt & skt $\Rightarrow$ cli &  pnt $\Rightarrow$ skt & pnt $\Rightarrow$ rel & pnt $\Rightarrow$ qdr & qdr $\Rightarrow$ skt & qdr $\Rightarrow$ rel & qdr $\Rightarrow$ cli\\
%&&&1/5-shot  &1/5-shot  &1/5-shot  &1/5-shot\\
\hline
\multicolumn{10}{c}{\textbf{5-way, 1-shot (acc/var)}} \\
\hline
 % MELDA & 32.29 & 41.79 & 33.11& 44.84 & 39.44 & 50.65 & 41.57 & 55.8 \\
IMSE & 59.91/0.014 & 80.14/0.015 & 87.31/0.017 & 55.39/0.018 & 80.26/0.019 & 88.75/0.017 & 69.12/0.019 & 76.27/0.015 & 73.64/0.018 \\ 
TSECS & 66.75/0.011 & 86.15/0.014 & 92.27/0.011 & 62.85/0.014 & 85.63/0.015 & 92.76/0.012 & 74.53/0.016 & 80.76/0.011 & 80.71/0.012 \\ 
E$^2$MPL & \textbf{72.54/0.009} & \textbf{87.20/0.012} & \textbf{93.77/0.009} & \textbf{69.97/0.012} & \textbf{88.31/0.013} & \textbf{93.31/0.009} & \textbf{79.96/0.015} & \textbf{84.19/0.009} & \textbf{83.42/0.003} \\
\hline \hline
\multicolumn{10}{c}{\textbf{5-way, 5-shot (acc/var)}} \\
\hline
IMSE & 63.81/0.013 & 82.36/0.010 & 89.62/0.011 & 61.25/0.018 & 82.86/0.011 & 87.93/0.012 & 71.10/0.013 & 78.73/0.006 & 76.82/0.014 \\
TSECS & 67.37/0.012 & 88.24/0.008 & 93.25/0.006 & 65.19/0.010 & 89.17/0.007 & 94.94/0.011 & 78.34/0.010 & 84.89/0.005 & 83.21/0.011 \\ 
E$^2$MPL & \textbf{75.14}/\textbf{0.011} & \textbf{89.53/0.005} & \textbf{94.58/0.001} & \textbf{78.59/0.004} & \textbf{92.81/0.002} & \textbf{96.83/0.011} & \textbf{83.89/0.008} & \textbf{88.52/0.003} & \textbf{87.64/0.007} \\
\hline
\end{tabular}
%\vspace{-0.3cm}
\end{table*}

% \textbf{Domain discriminator.} 
% To investigate the effect of leveraging ridge regression as domain discriminator, we replace it with three fully-connected layers (namely MELDA-FD) that need multiple steps of gradient update to generate a local optimal solution. We evaluate its performance and testing time for two feature embedding backbones (`96-192-384-512' and ResNet-12) in Table \ref{tab:discriminators}. Clearly, our MELDA using ridge regression performs better and takes much less time than MELDA-FD. \re{Also, our MELDA-$\theta_p$ has the best performance and its time cost is slightly higher than MELDA.} This validates the efficacy of our discriminator to address FS-UDA. Moreover, the comparison between two backbones shows that our MELDA becomes more effective when using the deeper model ResNet-12 to generate informative features, although incurring more time than using `96-192-384-512'. 

\textbf{Comparison of prompt learning methods.} 
To fully demonstrate the advantage of the designed prompts in our current work, we investigated our method with some previous prompt learning methods, including VPT \cite{vpt}, PGN \cite{PGN}, DAM-VP \cite{DAM-VP} and ProMetaR \cite{ProMetaR}. Given that these methods cannot be used directly in our setting, we have to put their prompt approach in our E$^2$MLP meta-prompt learning framework, substituting for our domain-shared prompts and task-specific prompts. The classification accuracies obtained in 5-way 1-shot and 5-way 5-shot setting are shown in \textbf{Table \ref{tab:tablet}}. As can be seen, our method achieves the best performance, which indicates the efficacy of our designed domain-shared prompts and task-specific prompts.

\textbf{Ablation study of the prompt components.} To better demonstrate the contributions of each component of our prompts, we evaluated the performance of each prompt component in Table \ref{tab:tablep}, where \textbf{all (-skt) $\Rightarrow$ skt} indicates the average accuracy when other domains except for \emph{sketch} are the source domain and \emph{sketch} is the target domain. The first line of Table \ref{tab:tablep} is the baseline, which is to freeze the CLIP and not to use any prompts. The second line uses only domain-shared prompts, which helps the model learn a more generalized representation of features, thereby mitigating domain gaps. The third line is the use of task-specific prompts, which are obtained by passing image data through a ResNet-12 network pre-trained with \emph{miniImageNet}. Unlike domain-shared prompts, this method is equivalent to introducing some prior knowledge of the current data, thus enhancing the feature extraction capability of the model. The fourth line is the effect of our combination of domain-shared and task-specific prompts, showing that the two prompts can complement each other and greatly improve the performance of the model.

% \textbf{Ablation study of three loss terms.} To better demonstrate our method, we investigate the three losses in Eqn. (\ref{e2}). %and the automatic update of parameters $\gamma_\omega,\gamma_d$ in Eqns.(\ref{e6}-\ref{e7}) during meta learning. 
% The resulting classification accuracy 
% %using backbone '96-192-384-512' 
% is shown in Table \ref{tab:table3}. The \textbf{all (-skt) $\Rightarrow$ skt} means the same as above. %represents the average accuracy of other domains except sketch as the source domain and sketch as the target domain.%\re{$\gamma_d^*/\gamma_\omega^*$ means the method automatically updating $\gamma_d$ and $\gamma_\omega$ with meta learner to achieve the optimum.} 

\begin{figure}[htbp]
\centering
%\vspace{-0.2cm}
\includegraphics[width=\columnwidth]{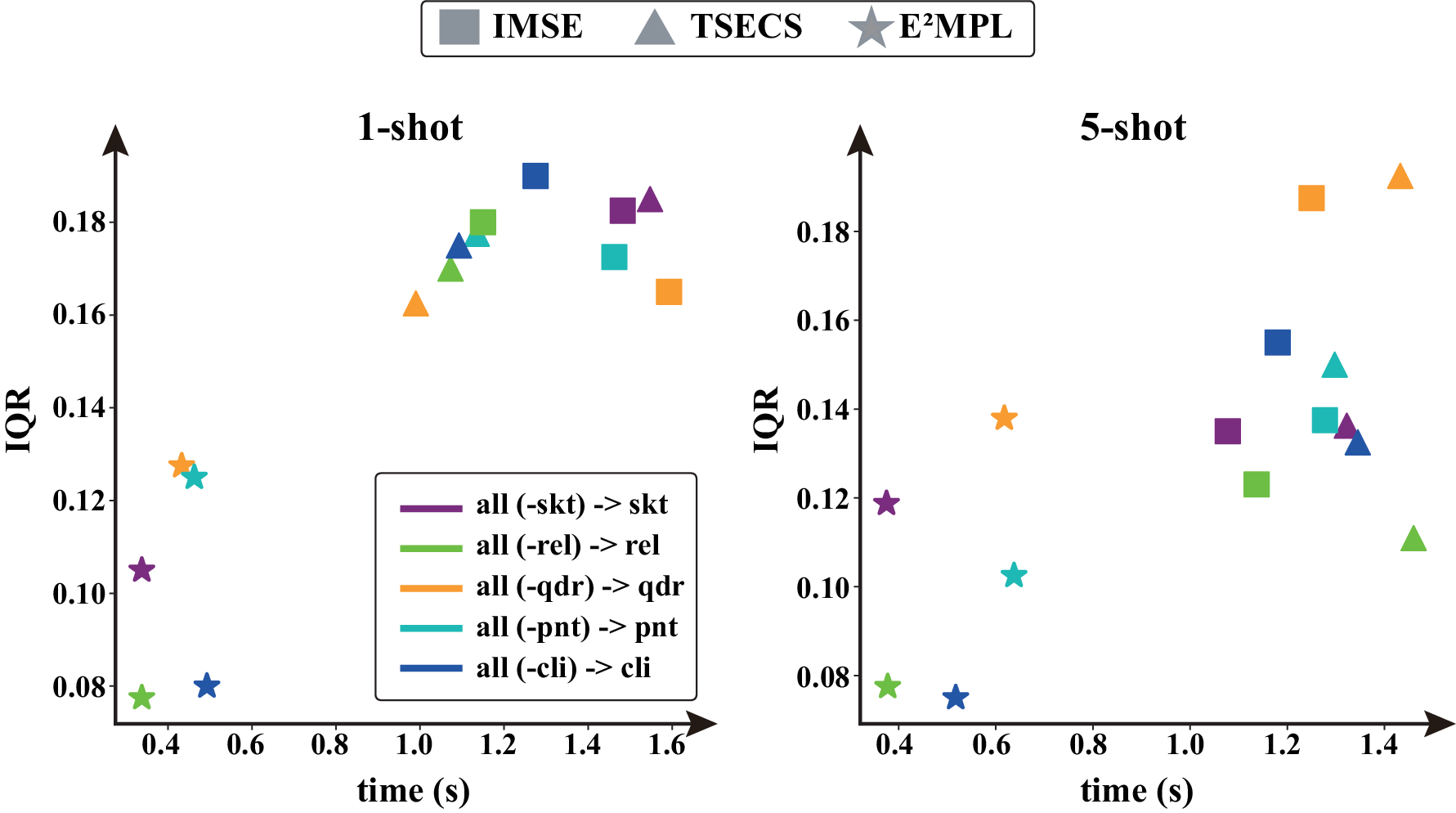}
%\vspace{-0.3cm}
%\\setlength{\abovecaptionskip}{2pt}
%\setlength{\belowcaptionskip}{-2pt}
\caption{Comparison of the average adaptation time (seconds) each task and classification accuracy change (boxplot IQR value) for 3600 test tasks. The proposed E$^2$MPL has lower time cost and a lower IQR value for both 5-way 1-shot and 5-way 5-shot training settings.} 
\label{fig3}
\end{figure}

\begin{figure*}[htbp]
\centering
\includegraphics[width=1.9\columnwidth]{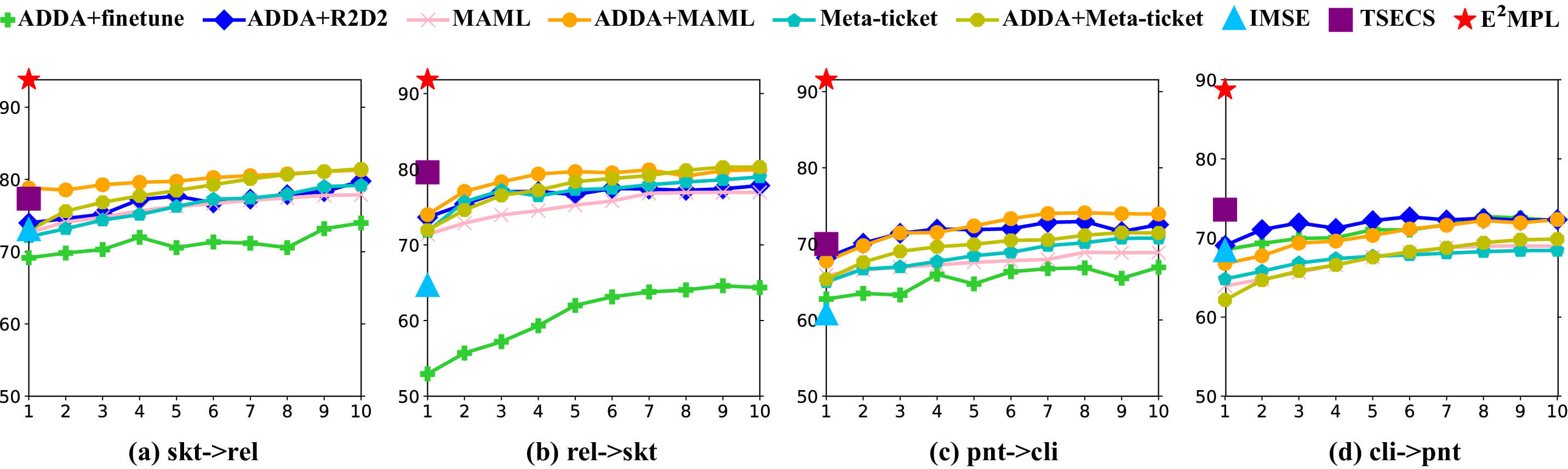}%{fig_compare_v2.eps}
\caption{Comparison of different methods for adapting to 3600 new 5-way 1-shot UDA tasks. Our E$^2$MPL is denoted as a red star and a green square respectively.}\label{fig5}
\end{figure*}

\subsection{Evaluation of Model Adaptation}
\textbf{Evaluation of adaptation time and stability to new tasks.} 
To validate the enduring and efficient performance of our method in new FS-UDA tasks, we evaluated the average time of model adaptation to each task, as well as the IQR value of the classification accuracy boxplot for 3600 test tasks, among the three FS-UDA methods (E$^2$MPL, IMSE, and TSECS).
Their time cost (in seconds) and accuracy IQR values are depicted in \textbf{Fig. \ref{fig3}}, where CLIP is the feature embedding backbone we used. Clearly, our E$^2$MPL using bilevel optimization meta-prompt learning takes much less time and lower IQR values than IMSE and TSECS. This validates the efficiency and stability of our meta-prompt learning framework to address FS-UDA.

\textbf{Model adaptation to new tasks.} To further show the performance of our method to adapt to new tasks by one-step update, we compare it with the eight baselines using meta-learning for adaptation (ADDA+finetune, ADDA+R2D2, MAML, ADDA+MAML, Meta-ticket, ADDA+Meta-ticket, IMSE, and TSECS), using the CLIP backbone network for all models. 
The first six methods are based on meta-learning following the setting in \cite{DBLP:conf/icml/FinnAL17}, and allow 10 gradient updates to adapt to new tasks, while the latter two methods are metric-based, classifying by directly measuring sample similarities with the support set without relying on the gradient optimization process.
Thus,  \textbf{Fig. \ref{fig5}} shows the nine methods in terms of the accuracy of one-step update or accuracy changing curves under multiple gradient updates. It is evident that the inclusion of the prompt module enables E$^2$MPL to outperform traditional methods that combine FSL and UDA. Additionally, although IMSE and TSECS have demonstrated strong capabilities, their performance remains significantly lower than E$^2$MPL because our model designs a lightweight but effective meta prompt learning framework.

\textbf{Domain Generalization.} To thoroughly understand the impact and reliability of the prompt module in domain generalization, we investigated the performance of model adaptation to a new target domain, which the model had not encountered during meta-training. We evaluated the averaged accuracy and variance values in 3600 new test tasks using CLIP as the backbone for 5-way 1-shot and 5-shot UDA tasks, shown in \textbf{Table \ref{tab:table5}}. It is evidently shown that our E$^2$MPL, with its designed meta-prompt learning, has better advantages in domain generalization and exhibits greater stability across different tasks.

\begin{table*}
\centering
%\small
%\scriptsize
%\vspace{-0.2cm}
\caption{Ablation study of the effect of the three losses in our E$^2$MPL (backbone CLIP), where \textbf{all (-skt) $\Rightarrow$ skt} indicates the average accuracy when other domains except for \emph{sketch} are the source domain and \emph{sketch} is the target domain.} %`96-192-384-512')} %varying combinations are evaluated on the testing sets of \emph{miniImageNet} (left) and \emph{Office-Home} (bottom). 
%The averaged accuracy values of target domain over 100 tasks under the 5-way, 1-shot or 5-shot setting are reported in percent, and the best performances are in bold. }
%\vspace{-0.1cm}
\label{tab:table3}
\renewcommand{\arraystretch}{1.2}
\setlength{\tabcolsep}{5pt}

\begin{tabular}{ccc|cccccccccccccc}
\hline
\multirow{2}{*}{$\mathcal{L}_{c}^{\text{meta}}$} & \multirow{2}{*}{$\mathcal{L}_{d}^{\text{meta}}$} & \multirow{2}{*}{$\mathcal{L}_{F}^{\text{meta}}$} %\multirow{2}{*}{$\gamma_d^{*}/\gamma_\omega^{*}$} & 
% \multicolumn{2a}{c|}{\emph{miniImageNet*}} &
% \multicolumn{2}{c}{\emph{Office-Home}} \\\cline{3-6}%\cline{7-8}
% &&\textbf{Ph $\Rightarrow$ Sk} &
% \textbf{Sk $\Rightarrow$ Ph} &
% \textbf{Cl $\Rightarrow$ Pr} &
% \textbf{Pr $\Rightarrow$ Cl} \\
& \multicolumn{2}{c}{\textbf{all(-skt) $\Rightarrow$ skt}} && \multicolumn{2}{c}{\textbf{all(-rel) $\Rightarrow$ rel}}&&\multicolumn{2}{c}{\textbf{all(-qdr) $\Rightarrow$ qdr}} && \multicolumn{2}{c}{\textbf{all(-pnt) $\Rightarrow$ pnt}} && \multicolumn{2}{c}{\textbf{all(-cli) $\Rightarrow$ cli}}\\\cline{4-5}\cline{7-8}\cline{10-11}\cline{13-14}\cline{16-17}
&&& 1-shot&5-shot&&1-shot&5-shot&&1-shot&5-shot&&1-shot&5-shot&&1-shot&5-shot\\
%&&&1/5-shot  &1/5-shot  &1/5-shot  &1/5-shot\\
\hline
\checkmark & & & 83.58&88.29   && 86.83&93.55   & & 71.34&76.20   && 79.71&84.27  && 86.73&91.69   \\
 & \checkmark &&   82.63&86.59   && 87.45&91.26  && 69.10&75.35   && 79.30&82.55  && 85.41 &89.64   \\
& & \checkmark &   67.16&71.54   && 73.86&76.15  && 56.10&63.21   && 65.41&70.27  && 71.70 &74.67   \\
%\checkmark & \checkmark &  &  33.77/41.05   & 34.85/42.73     & 32.86/42.08  & \textbf{34.15}/40.14  \\
\checkmark & \checkmark&& 88.05&92.83 &&92.03& 96.53 &&76.57&85.17&&84.38&89.80&& 91.03&95.02\\
\checkmark & \checkmark& \checkmark& \textbf{89.07}&\textbf{93.29}&&\textbf{93.26}&\textbf{96.89}&&\textbf{78.49}&\textbf{85.31}&&\textbf{85.60}&\textbf{90.03}&& \textbf{92.60}&\textbf{95.69}\\
\hline
\end{tabular}
%\vspace{-0.3cm}
\end{table*}

\subsection{More Results}

\textbf{Ablation study of three loss terms.} To better demonstrate our method, we investigate the impact of the three losses in Eqn. (\ref{e6}). The resulting classification accuracies are shown in \textbf{Table \ref{tab:table3}}. The \textbf{all (-skt) $\Rightarrow$ skt} means the same as above. The first line $\mathcal{L}_c^{\text{meta}}$ means that the model tries to adapt to categories without considering domain adaptation. The second line $\mathcal{L}_{d}^{\text{meta}}$ means that the model tries to transfer from the source domain to the target domain without considering classification generalization. $\mathcal{L}_{f}^{\text{meta}}$ means that only the semantic discriminative features of the classes are considered. Comparison of these three lines with the last line (\emph{i.e.,} our E$^2$MPL) indicates that using $\mathcal{L}_c^{\text{meta}}$, $\mathcal{L}_{d}^{\text{meta}}$ and $\mathcal{L}_{f}^{\text{meta}}$ are more effective than only using one of them. This shows the necessity of E$^2$MPL to jointly perform classification generalization and domain adaptation during meta-training for the FS-UDA setting.  %\re{Moreover, compared with the first three rows setting a fixed value (\emph{i.e.,} 1) for $\gamma_\omega$ and $\gamma_d$, our method (at last) in Table \ref{tab:table3} updates the parameters along with learning the meta learner, showing the efficacy of using the meta learning strategy to learn regularization parameters.}
%\re{\textbf{In addition, parameter analysis and more results can be found in the supplementary material.}}

% \textbf{Ablation study of prompt learning.} 
% In order to fully know the role of prompt learning in our current work, we investigated the various components in the prompt module. The first part represents learning without prompts, using only the frozen CLIP as the backbone. The second part is using simple input prompts. The third part is the use of domain-shared prompts, that is, to make a projection of all prompts to enhance the relationship between prompts. The fourth part is to use the task-specific prompts based on the domain sharing prompt, that is, the prompts generated by the images passing through the prompt network. Task-specific prompts provide some prior knowledge for prompt learning.

\begin{table}
\centering
%\small
%\scriptsize
\caption{Ablation study of the effect of metric method and similarity matrix in our E$^2$MPL (backbone CLIP)} %varying combinations are evaluated on the testing sets of \emph{miniImageNet} (left) and \emph{Office-Home} (bottom). MELDA-$\theta_p$
%The averaged accuracy values of target domain over 100 tasks under the 5-way, 1-shot or 5-shot setting are reported in percent, and the best performances are in bold. }
\vspace{-0.1cm}
\label{tab:table4}
\setlength{\tabcolsep}{6pt}
\begin{tabular}{ccc|ccccc}
\hline
 & \multirow{2}{*}{$\gamma_p$} & \multirow{2}{*}{\textbf{sinkhorn}} & %\multirow{2}{*}{$\gamma_d^{*}/\gamma_\omega^{*}$} & 
% \multicolumn{2}{c|}{\emph{miniImageNet*}} &
% \multicolumn{2}{c}{\emph{Office-Home}} \\\cline{3-6}%\cline{7-8}
% &&\textbf{Ph $\Rightarrow$ Sk} &
% \textbf{Sk $\Rightarrow$ Ph} &
% \textbf{Cl $\Rightarrow$ Pr} &
% \textbf{Pr $\Rightarrow$ Cl} \\
\multicolumn{2}{c}{\textbf{skt $\Rightarrow$ rel}} && \multicolumn{2}{c}{\textbf{rel $\Rightarrow$ skt}}\\\cline{4-5}\cline{7-8}
& & &1-shot&5-shot&&1-shot&5-shot\\
%&&&1/5-shot  &1/5-shot  &1/5-shot  &1/5-shot\\
\hline
 \multirow{4}{*}{\textbf{Euc}} & & & 89.41&94.48   && 88.03&89.61 \\
&  & \checkmark &  92.82&95.60   && 90.81&91.55 \\
& \checkmark & &  91.81&95.04   && 90.15&92.85 \\
& \checkmark &\checkmark &\textbf{94.80}& \textbf{98.19}   && \textbf{92.83}&\textbf{94.76} \\
\hline
 \multirow{4}{*}{\textbf{Cos}} & & & 89.91&91.39   && 87.28&88.07 \\
 & & \checkmark & 92.56& 94.28 && 89.58&90.44 \\
 & \checkmark &  &91.66&95.37 && 89.49&90.56 \\
 & \checkmark & \checkmark& 93.56& 97.73 && 91.94&93.16 \\
%\checkmark & \checkmark &  &  33.77/41.05   & 34.85/42.73     & 32.86/42.08  & \textbf{34.15}/40.14  \\
\hline
\end{tabular}
%\vspace{-0.3cm}
\end{table}

\textbf{Ablation study of the domain adapter. } To better understand the domain adapter, we investigate the following three aspects: 1) using two different methods (Euclidean distance and Cosine similarity) to calculate the similarity matrix $A$ in formula (\ref{e11}); 2) whether to use a learnable regularization coefficient $\gamma_p$; 3) whether to regularize the similarity matrix using sinkhorn. The experimental results from \emph{sketch} to \emph{real} and from \emph{real} to \emph{sketch} are shown in \textbf{Table \ref{tab:table4}}. Obviously, the accuracy of using Euclidean distance is better than that of using Cosine similarity. Moreover, both the learnable regularization coefficients and adjacency matrix regularization can improve the performance of the domain adapter.

\begin{figure}%{r}{0.32\textwidth}
\centering
\includegraphics[width=\columnwidth]{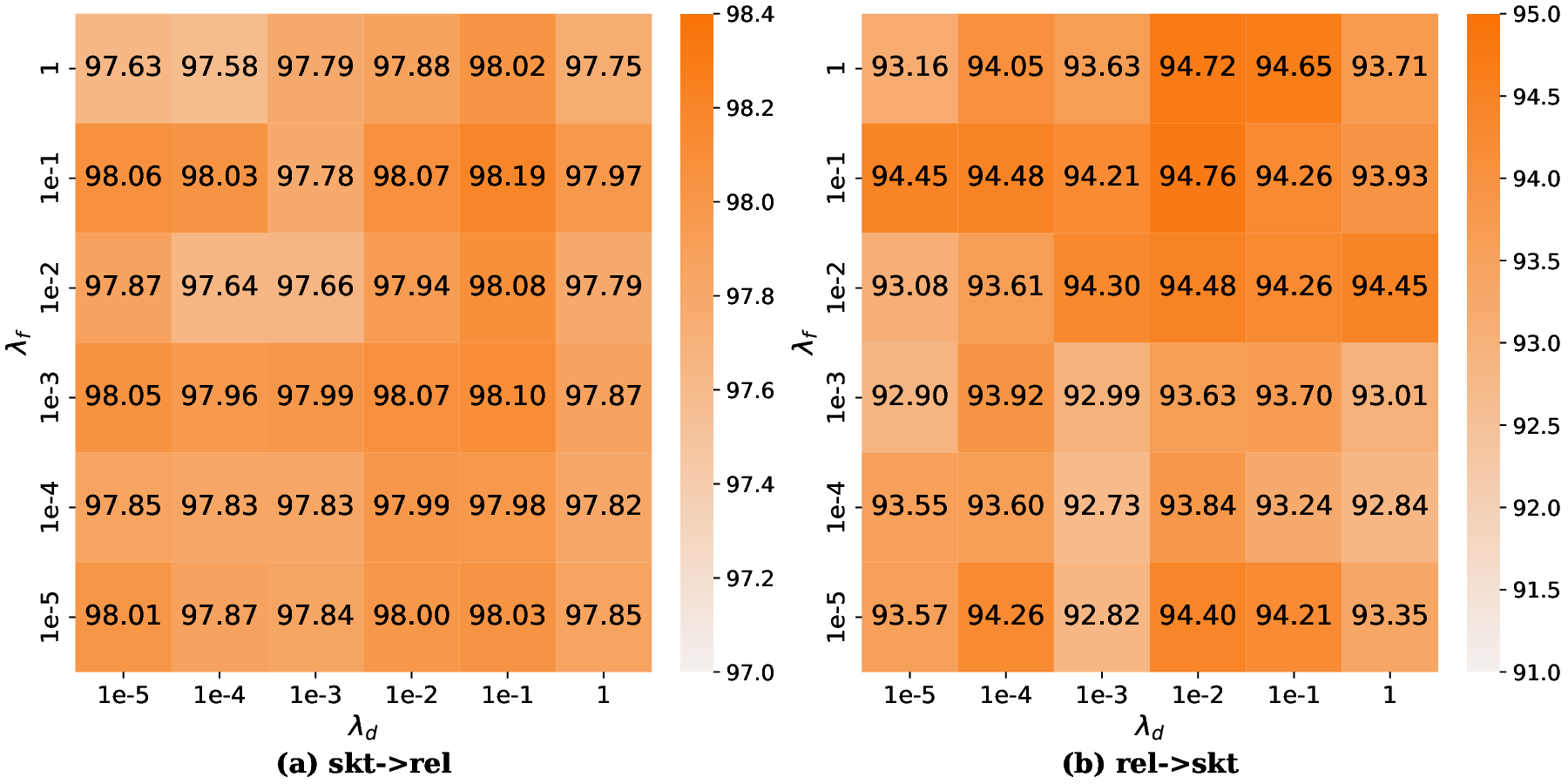}
\caption{The classification accuracy (\%) on \emph{DomainNet} with parameters $\lambda_d$ and $\lambda_f$ (backbone CLIP).}
\label{fig_heat}
%\vspace{-0.3cm}
\end{figure}

\begin{figure}[htbp]
\centering
\subfigure[skt$\leftrightarrow$ rel]{ % miniImageNet*
\includegraphics[width=0.46\columnwidth]{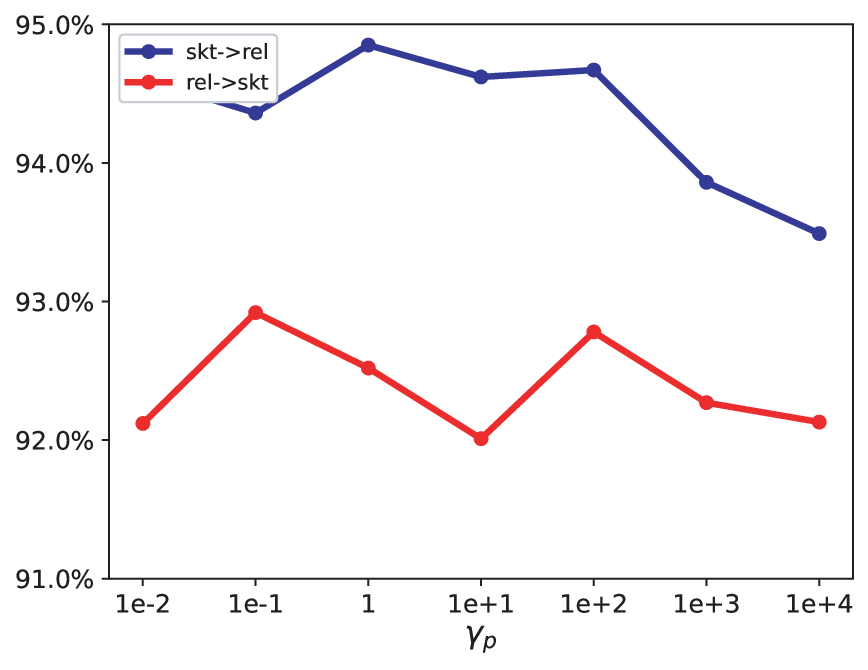}}
\subfigure[pnt$\leftrightarrow$cli]{ % Office-Home
\includegraphics[width=0.46\columnwidth]{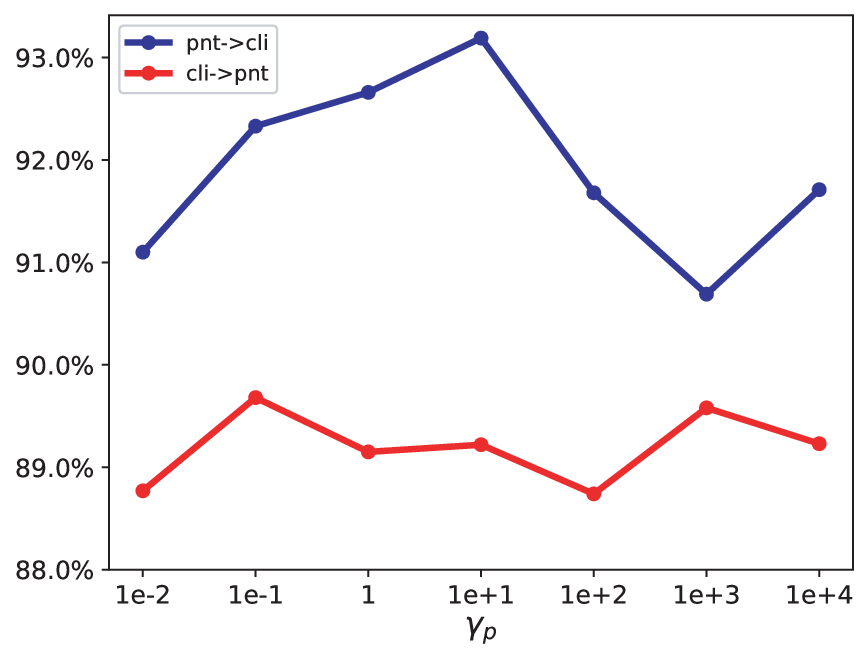}}
%\vspace{-0.4cm}
\caption{The classification accuracy (\%) on both datasets with varying parameter $\gamma_p$ (backbone CLIP)}\label{fig_gamma}
\end{figure}

\textbf{Effect of parameters $\lambda_d$ and $\lambda_f$.} We evaluate the impact of parameters $\lambda_d$ and $\lambda_f$ in Eqn. (\ref{e6}) on the performance of FS-UDA with 5-way $k$-shot setting for \emph{DomainNet}, shown in Fig. \ref{fig_heat}. According to Eqn. (\ref{e6}), $\lambda_d$ and $\lambda_f$ are used to balance the classification and domain adaptation, and the larger $\lambda_d$ will enhance the domain adaptation of the model, while the larger $\lambda_f$ will make the learned features more discriminative. We perform a grid search of $\lambda_d$ and $\lambda_f$ within $\{10^{\text{-5}},10^{\text{-4}},...,1\}$. As can be seen in Fig. \ref{fig_heat}, when the value of $\lambda_d$ is about $10^{\text{-2}}$ to $10^{\text{-1}}$ and the value of $\lambda_f$ is $10^{\text{-1}}$ for \emph{DomainNet}, the optimal results are obtained in the validation set. 
\textbf{Effect of parameter $\gamma_p$.} We show the effect of different $\gamma_p$ values on the model performance through line plots. $\gamma_p$ represents the regularization coefficient in Eqn. (\ref{e9}). \textbf{Fig. \ref{fig_gamma}} shows the classification accuracy, as $\gamma_p$ varies within $\{10^{\text{-2}}, 10^{\text{-1}},...,10^{\text{4}}\}$. The optimal $\gamma_p$ value is set as $10^{\text{3}}$ for \emph{DomainNet}. This indicates that regularization of $\theta_\mathcal{T}$ is effective for domain adaptation. % Likely, we evaluate the effect of $\gamma_p$ by using the same setting as the above. \ywq{$\gamma_p$ represents the regularization term coefficient in Eqn. (\ref{e9}) .} Fig. \ref{fig_gamma} shows the classification accuracy, as $\gamma_p$ varies within $\{10^{\text{-2}}, 10^{\text{-1}},...,10^{\text{4}}\}$. The optimal $\gamma_p$ value is set as $10^{\text{3}}$ for \re{\emph{DomainNet}}. This indicates that the regularization of $\theta_\mathcal{T}$ is effective for domain adaptation.

\section{Conclusion}
This work focuses on a realistic problem setting of FS-UDA, and for considering the stability and efficacy, we propose an enduring and efficient meta prompt learning framework (E$^2$MPL), and design domain-shared prompts that learn domain-invariant knowledge to mitigate domain gaps, as well as task-specific prompts that generate the task-dependent prompts from pretrained prompt network to quickly adapt to new tasks. We design the bilevel optimization framework for meta-prompt learning, and leverages ridge regression and domain projection with closed-form solutions, avoiding multi-step updates in meta-training, so the meta model can be efficiently trained and adapted. Extensive experiments have confirmed the efficacy of E$^2$MPL. In the further work, we will explore the more effective method for FS-UDA setting in real applications, \emph{e.g.,} where the number of samples in each class may be varied.
\bibliographystyle{IEEEtran}
\bibliography{IEEEabrv,egbib}
\end{document}